\documentclass[letterpaper]{article} 
\usepackage{aaai25}  
\usepackage{times}  
\usepackage{helvet}  
\usepackage{courier}  
\usepackage[hyphens]{url}  
\usepackage{graphicx} 
\usepackage{natbib}  
\usepackage{caption} 
\usepackage{algorithm}
\usepackage{algorithmic}

\usepackage{url}
\usepackage{times}
\usepackage{epsfig}
\usepackage{amsmath}
\usepackage{amssymb}
\usepackage{subcaption}
\usepackage{multirow}
\usepackage[flushleft]{threeparttable}
\usepackage{pifont}

\usepackage{booktabs}

\newlength\savewidth\newcommand\shline{\noalign{\global\savewidth\arrayrulewidth\global\arrayrulewidth 1pt}\hline\noalign{\global\arrayrulewidth\savewidth}}
\newcommand{\tablestyle}[2]{\setlength{\tabcolsep}{#1}\renewcommand{\arraystretch}{#2}\centering\footnotesize}

\usepackage{color}
\usepackage{xcolor}         
\usepackage{array}
\usepackage{makecell}
\usepackage{tabularray}
\usepackage{colortbl}

\usepackage{arydshln}
\usepackage{array}

\definecolor{mygray}{gray}{.9}

\usepackage{newfloat}
\usepackage{listings}
\DeclareCaptionStyle{ruled}{labelfont=normalfont,labelsep=colon,strut=off} 
\floatstyle{ruled}
\newfloat{listing}{tb}{lst}{}
\floatname{listing}{Listing}
\title{Stable Mean Teacher for Semi-supervised Video Action Detection}
\author{
    Akash Kumar, Sirshapan Mitra, Yogesh Singh Rawat\\
}
\affiliations{
    University of Central Florida\\
    \{akash.kumar, sirshapan.mitra, yogesh\}@ucf.edu \\
    Project Page: \url{https://akash2907.github.io/smt_webpage}
}

\usepackage{bibentry}

\begin{document}

\maketitle 
\begin{abstract}
  In this work, we focus on semi-supervised learning for video action detection. 
Video action detection requires spatio-temporal localization in addition to classification and limited amount of labels make the model prone to unreliable predictions.
We present \textbf{\textit{Stable Mean Teacher}},
a simple end-to-end student-teacher based framework 
which benefits from \textit{\textbf{improved}} and \textit{\textbf{temporally consistent}} pseudo labels.
It relies on a novel \textit{\textbf{ErrOr Recovery (EoR)}} module
which learns from students' mistakes on labeled samples and transfer this to the teacher to improve psuedo-labels for unlabeled samples.
Moreover, existing spatio-temporal losses does not take temporal coherency into account and are prone to temporal inconsistencies. To overcome this, we present \textbf{\textit{Difference of Pixels (DoP)}}, a  simple and novel constraint focused on temporal consistency which leads to coherent temporal detections.
We evaluate our approach on four different spatio-temporal detection benchmarks, UCF101-24, JHMDB21, AVA and Youtube-VOS. Our approach outperforms the supervised baselines for action detection by an average margin of \textbf{23.5\%} on UCF101-24, \textbf{16\%} on JHMDB21, and, \textbf{3.3\%} on AVA. Using merely 10\% and 20\% of data, it provides a competitive performance compared to the supervised baseline trained on 100\% annotations on UCF101-24 and JHMDB21 respectively.  We further evaluate its effectiveness on AVA for scaling to large-scale datasets and Youtube-VOS for video object segmentation demonstrating its \textit{\textbf{generalization capability}} to other tasks in the video domain. Code and models are publicly available at:
\end{abstract}

\begin{links}
\link{Code}{https://github.com/AKASH2907/stable-mean-teacher}
\link{Models}{https://huggingface.co/akashkumar29/stable-mean-teacher}
\end{links}

\section{Introduction}
\label{sec:intro}

Video action detection is a challenging problem with several real-world applications in security, assistive living, robotics, and autonomous-driving. What makes the task of video action detection challenging is the requirement of spatio-temporal localization in addition to video-level activity classification. This requires annotations on each video frame, which can be cost and time intensive. In this work, we focus on semi-supervised learning (SSL) to develop label efficient method for video action detection.

\begin{figure*}[t!]
\begin{center}
    \includegraphics[width=\textwidth]{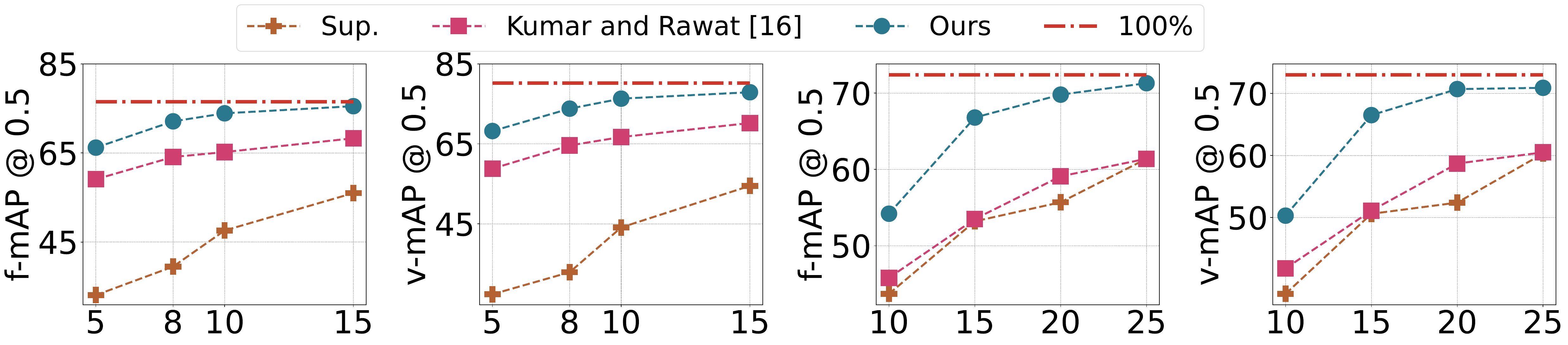}
\end{center}

\caption{ \textbf{Performance overview:} Stable Mean Teacher provides comparable performance with 10\% (\textbf{UCF101-24}; left two plots) and 20\% (\textbf{JHMDB-21}; right two plots) labels when compared with fully supervised approach which is trained on 100\% annotations. It consistently outperforms existing state-of-the-art \shortcite{Kumar_2022_CVPR} and supervised baseline on both f-mAP and v-mAP with good margin on both UCF101-24 and JHMDB-21 at all different percentages of labeled set. \textit{x-axis} shows annotation percentage in each plot. 
}
 
\label{fvmap}
\end{figure*}

Semi-supervised learning (SSL) is an active research area with two prominent approaches: iterative proxy-label \cite{rizve2020defense} and consistency based methods \cite{mt}. Iterative proxy-label methods, although effective, are not suitable for the video domain due to their lengthy training cycles. On the other hand, consistency-based approaches offer end-to-end solutions, requiring only a single pass through the dataset for training. While most of the existing research in this area has focused on image classification \cite{const1, mt, const3, const4} and object detection 
\cite{semiobj1, semiobj5, semiobj6},
limited efforts have been made in the video domain with works only focusing on classification \cite{semivideocls1, semivideocls4, kumar2023benchmarking}.
We also observe that Mean Teacher \cite{mt} based approaches have demonstrated superior performance among consistency-based methods. Building upon the success of student-teacher learning in image domain, we extend it to video domain for spatio-temporal detection tasks.  

Video action detection, in contrast to classification and object detection, poses additional challenges for semi-supervised learning. It is a complex task that combines both classification and spatio-temporal localization which suffers performance degradation under limited availability of labels. 
Moreover, the detections have to be temporally coherent in addition to spatial correctness. 
Therefore, it is challenging to generate high-quality spatio-temporal pseudo-labels for videos.  To overcome these challenges, we propose \textbf{\textit{Stable Mean Teacher}}, a simple end-to-end framework. It is an adaptation of Mean Teacher where we study both {\textit{classification}} and {\textit{spatio-temporal consistencies}} to effectively utilize the pseudo-labels generated for unlabeled videos.

Stable Mean Teacher consists of a novel \textbf{\textit{ErrOr Recovery (EoR)}} module which learns from the student's mistakes on labeled samples and transfer this learning to the teacher for improving the spatio-temporal pseudo-labels generated on unlabeled set. EoR improves pseudo labels, but ignore temporal coherency which is important for action detection. To overcome this, we introduce \textbf{\textit{Difference of Pixels (DoP)}}, a simple and novel constraint that focuses on the temporal coherence and helps in generating consistent spatio-temporal pseudo-labels from unlabeled samples.


In summary, we make the following contributions:
\begin{itemize}
    \item 
    We propose \textbf{\textit{Stable Mean Teacher}}, a simple end-to-end approach for semi-supervised video action detection. 
    \item We propose a novel \textbf{\textit{ErrOr Recovery (EoR)}} module, which learns from the student's mistakes and helps the teacher in providing a better supervisory signal under limited labeled samples. 
    \item We propose \textbf{\textit{Difference of Pixels (DoP)}}, a simple and novel constraint which focuses on temporal consistencies and leads to coherent spatio-temporal predictions. 
\end{itemize}

We perform a comprehensive evaluation on three different action detection benchmarks. Our study demonstrates significant improvement over supervised baselines, consistently outperforming the state-of-the-art approach for action detection (Figure \ref{fvmap}). We also demonstrate the generalization capability of our approach on video object segmentation.

\section{Related Work}

\noindent
\textbf{Video action detection}
Video action detection comprises two tasks: action classification and spatio-temporal localization. Some of the initial attempts to solve this problem are based image-based object detectors such as RCNN \cite{frcnn} and DETR \cite{detr}, where detection at frame-level is used for video-level activity classification \cite{yang2019step, hou2017tcnn, yang2017spatio, Dave_2022_WACV, Zhao_2022_CVPR, evad, bmvit}.  Most approaches involve two-stages, where localization is performed using a region proposal network which is classified into activities in the second stage \cite{gkioxari2017detecting, yang2019step, hou2017tcnn, yang2017spatio}. Recently, some encoder-decoder based approaches have been developed on CNN \cite{duarte2018videocapsulenet} and transformer-based \cite{Zhao_2022_CVPR, Wu2023STMixerAO, evad, bmvit} backbones which simplify the two-stage video action detection process. However, transformer-based backbones are complex and heavy, involving multiple modules. In a recent work \cite{Kumar_2022_CVPR}, the authors further simplify VideoCapsuleNet \cite{duarte2018videocapsulenet} to reduce computation cost with minor performance trade-off. In this work, we make use of this optimized approach as our base model for video action detection. 

\noindent
\textbf{Weakly-supervised learning} 
Some recent works in weakly-supervised learning attempts to overcome the high labeling cost for action detection \shortcite{Escorcia2020GuessWA, Arnab2020UncertaintyAwareWS, weakly5, weakly4, cheron_2018, weakly6}. These approaches require either video-level annotations or annotations only on few frames. However, they rely on external detectors \cite{frcnn, ssd, detr} which introduces additional learning constraints. Even with the use of per-frame annotations along with video-level labels, the performance is far from satisfactory when compared with supervised baselines. 
In our work, we only use a subset of labeled videos that are fully annotated and demonstrate competitive performance when compared with supervised methods.

\noindent
\textbf{Semi-supervised learning} 
have shown great promise in label efficient learning. Most of the efforts are focused on classification tasks \cite{mt, Ke2019DualSB} where sample level annotation is required, such as object recognition \cite{semiobj6, semiobj4} and video classification \cite{semivideocls2, semivideocls4}. 
These efforts can be broadly categorized into iterative pseudo-labeling \cite{pseudocls} and consistency-based \cite{mixmatch, fixmatch} learning. Consistency-based approaches are efficient  as the learning is performed in a single step in contrast to several iterations in iterative pseudo-labeling \cite{rizve2020defense}. Mean teacher \cite{mt} is a strong consistency-based approach where the pseudo-labels generated by the teacher are used to train a student in both image classification \cite{mt} as well as object detection \cite{semiobj1, semiobj3, semiobj4, semiobj5, semiobj6, semiobj7, semiobj8}. In \cite{pham2021meta}, the authors proposed to utilize feedback from student for teacher based on meta-learning which requires two-step training with additional computation cost.

\begin{figure*}[t!]
\begin{center}
\includegraphics[width=\linewidth]{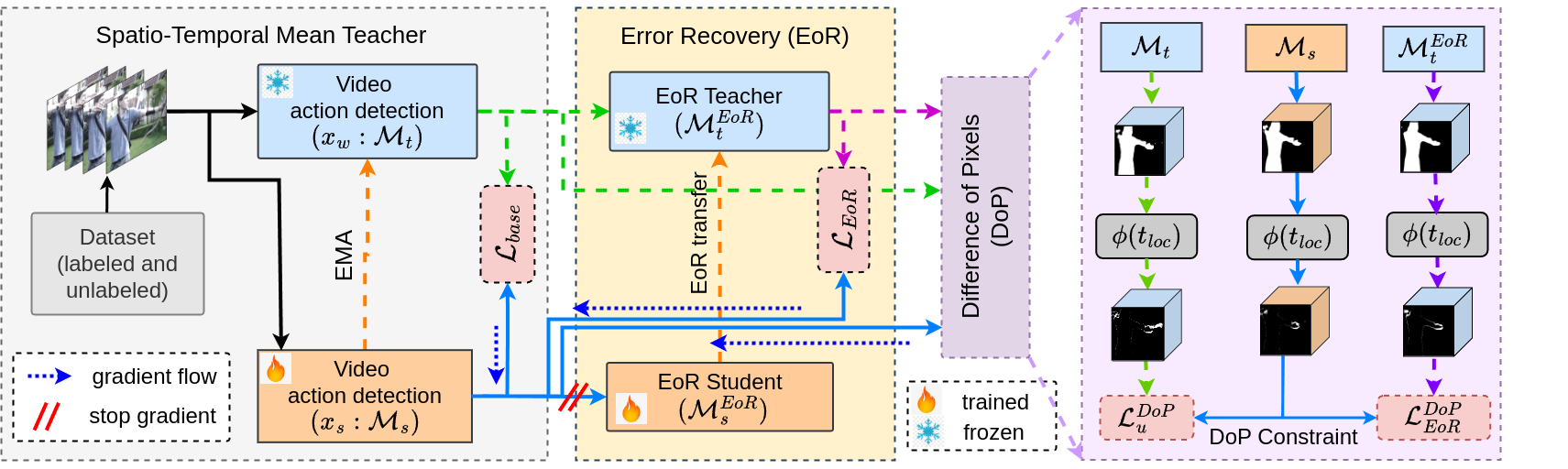}
\end{center}

\caption{ 
\textbf{Overview of Stable Mean Teacher.} The two key components to improve the quality of spatio-temporal pseudo label: 1) \textit{Error Recovery}: refines the spatial action boundary, 2) \textit{DoP constraint}: induces \textit{temporal coherency} on predicted spatio-temporal pseudo labels. 
}
 
\label{fig:main_arch}
\end{figure*}

Different from all these, we focus on videos where the temporal dimension adds more complexity to the problem. There are some recent works focusing on videos, but they are limited to video classification  \cite{semivideocls1,semivideocls2,semivideocls3, semivideocls4} and temporal action localization \cite{semitad1,semitad2,semitad3}  where per frame dense spatio-temporal annotations is not required. We focus on video action detection, which requires spatio-temporal localization on every frame of the video in addition to video level class predictions. More recently, a PI-based consistency approach \cite{Kumar_2022_CVPR, Singh_Rana_Kumar_Vyas_Rawat_2024} has been explored for semi-supervised video action detection. Different from this, we propose a Mean Teacher based approach adapted for video activity detection task which achieves better performance.

\section{Methodology}
\label{sec:method}
\noindent \textbf{Problem formulation}
Given a set of labeled samples $X_{L}: \{ x_{i}, y_{i}, f_{i} \}_{i=0}^{i=N_{l}}$ and an unlabeled subset $X_{U}:\{x_{i}\}_{i=0}^{i=N_{u}}$, where $x$ is a video and $y$ and $f$ corresponds to class label and frame level annotation with $N_l$ labeled and $N_u$ unlabeled samples. The labeled videos are annotated with a ground-truth class and frame-level spatio-temporal localization denoted as $y_{t}$ and $f_{t}$ respectively . Our goal is to train an action detection model $(M)$ using both labeled and unlabeled data.

\noindent \textbf{Overview} An overview of the proposed approach is illustrated in Figure \ref{fig:main_arch}. As shown in this Figure, Stable Mean Teacher follows a student-teacher approach adapted for video action detection task where the teacher model generates pseudo-labels using weak augmentations for the student who learns from these pseudo-labels on strongly augmented samples. 
Each video sample $(x_{i})$ is augmented to generate two views: strong $(x_{s})$ and a weak $(x_{w})$. We use the same action detection model $M$ as a teacher $(\mathcal{M}_{t})$ and as a student $(\mathcal{M}_{s})$. Each of these models have two outputs; action classification logits, $t_{cls}$ and $s_{cls}$, and raw spatio-temporal localization map, $t_{loc}$ and $s_{loc}$, respectively for teacher and student. 
To generate a better and confident spatio-temporal pseudo-label, the teacher learns from the student's mistakes on labeled samples to improve its pseudo-labels with the help of an \textit{Error Recovery (EoR)} module which is trained jointly. We pass $t_{loc}$ and $s_{loc}$ to Error Recovery module, $(\mathcal{M}^{EoR}_{t})$ and $(\mathcal{M}^{EoR}_{s})$, which generates refined localization maps, $t_{loc}^{EoR}$ and $s_{loc}^{EoR}$ respectively. To further induce temporal coherency in the predicted spatio-temporal pseudo label, we apply \textit{Difference of Pixels (DoP)} constraint for temporal refinement on the pseudo label for the student.

\noindent \textbf{Background}  We use Mean Teacher \shortcite{mt}, a student-teacher training scheme as our baseline approach. In Table \ref{tab:comparison_sota_ucf101}, we show that baseline Mean teacher works, however, it only exploits classification consistency, whereas video action detection task requires optimizing both classification and spatio-temporal localization task simultaneously. 

To address this issue, we \textit{adapt} \shortcite{mt} for action detection to formulate our base model with capability to generate spatio-temporal pseudo-labels required  for this task. Similar to Mean Teacher \shortcite{mt}, we use the teacher's prediction as a pseudo-label for the student model which attends to a strong perturbed version of the video. The teacher's model parameters $(\theta_{teacher})$ are updated via Exponential Moving Average (EMA) of the student's model parameters $(\theta_{student})$ with a decay rate of $\beta$. This update can be defined as, 
\begin{equation}
    \theta_{teacher} = \beta \theta_{teacher} + (1-\beta)\theta_{student}.
    \label{eq:ema_t}
\end{equation}
This base setup is trained using both classification and spatio-temporal loss and is defined as $\mathcal{L}_{base}$.
\begin{equation}
\begin{aligned}
    \mathcal{L}_{base}  &= \mathcal{L}_{base}^{cls} +   \mathcal{L}_{base}^{loc}
    \label{eq:lbase}
\end{aligned}
\end{equation}
where, $\mathcal{L}_{base}^{cls}$ represents classification loss and $\mathcal{L}_{base}^{loc}$ represents spatio-temporal localization loss. Moving forward, we refer STMT as \textit{the base model} in our work.
\subsection{Stable Mean Teacher}
\label{sec:smt}

\begin{figure*}[t!]
\begin{center}
\includegraphics[width=\linewidth]{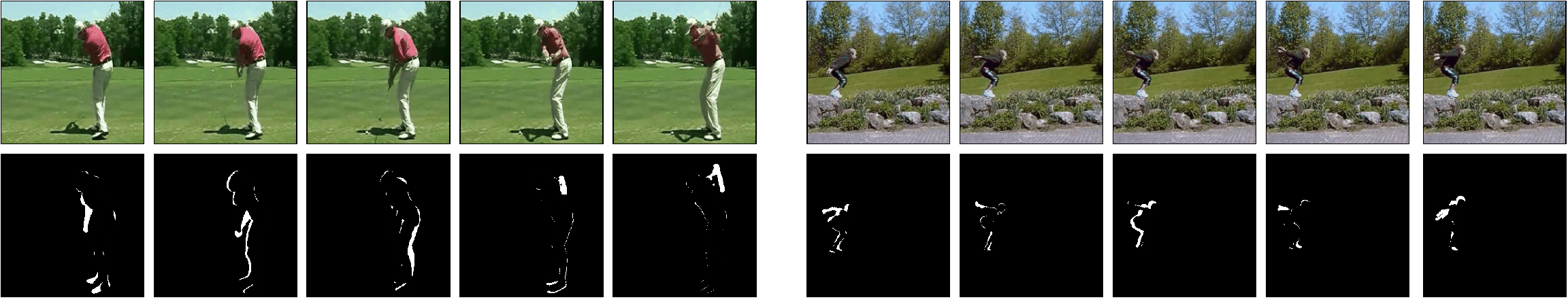}
\end{center}
\caption{ \textbf{Visualization of Difference of Pixels (DoP).} First row shows the RGB frames, second row shows the pixel difference map of ground truth along temporal dimension. We show two scenarios: \textit{Left:} Static: constant background; actor in motion, and \textit{Right:} Dynamic: changing background; actor in motion. Temporal difference emphasizes on the variation of boundary pixels between consecutive frames. 
}
 
\label{fig:dop}
\end{figure*}
The performance of the base model relies on the quality of the pseudo-labels generated by $(\mathcal{M}_{t})$. However, with limited labels, since the model focuses on two tasks: classification and localization simultaneously, it relies on the samples available per class. This limits the generalization capability of the model $(\mathcal{M}_{t})$ to generate high-quality pseudo-labels. To address this issue, we propose an Error Recovery module to improve the localization in a class-agnostic learning.

\subsubsection{Error Recovery (EoR)}
The Error Recovery module $(\mathcal{M}^{EoR})$ focus on correcting mistakes of the student model in spatio-temporal localization. These mistakes are approximated by an EoR module which attempts to recover these in class-agnostic learning. 
This is advantageous since model solely focuses  on localization task disregarding specific action class. 
This in turns  enriches the model's ability to localize actors accurately which potentially generates a better pseudo label for the student's model $(\mathcal{M}_{s})$. 
The base student model first tries to localize the actor and this is passed as input to the EoR module. The EoR module only focus on refining the localization without worrying about the type of activity. Therefore it can be trained in a class-agnostic manner.
EoR module is first trained to recover the students spatio-temporal mistakes on labeled samples with strong augmentations. Once trained, it is used to recover the mistakes on unlabeled samples with weak augmentations to improve the pseudo labels generated by the teacher. 
This in-turn improves pseudo labels for the student to learn from unlabeled samples.  
The model parameters $(\theta_{t}^{EoR})$ for $\mathcal{M}^{EoR}_{t}$ are updated via EMA of $\mathcal{M}^{EoR}_{s}$ parameters $(\theta_{s}^{EoR})$ with the same decay rate $\beta$ as describe in Eq. \ref{eq:ema_t}. The update is defined as,
\begin{equation}
    \theta_{t}^{EoR} = \beta \theta_{t}^{EoR} + (1-\beta)\theta_{s}^{EoR}.
    \label{eq:ema_aux}
\end{equation}
The Error Recovery module $\mathcal{M}^{EoR}_{s}$ does not use any pre-trained weights and is jointly trained with the base model on labeled samples in an end-to-end learning.
The student's prediction will be more distorted than the teacher's due to strong augmentations. This will help $\mathcal{M}^{EoR}_{t}$ to provide a more confident pseudo label on a weakly augmented sample. 
The loss $(\mathcal{L}_{EoR})$ is calculated between student's base model $(\mathcal{M}_{s})$ output and the refined pseudo label by $\mathcal{M}_{t}^{EoR}$, 
\begin{equation}
\mathcal{L}_{EoR}  =  MSE(\mathcal{M}_{t}^{EoR}(t_{loc}), s_{loc}) ,  
\label{eq:aux_loss}
\end{equation}
where $MSE$ is Mean-Squared-Error.

\subsubsection{Difference of Pixels (DoP)}
The Error Recovery module enhances the spatial localization of pseudo labels. However, in the context of videos, predictions need to maintain \textit{consistency} over time. To ensure this temporal coherency across frames, we introduce a novel training constraint named Difference of Pixels (DoP). This approach is motivated by the limitations of conventional loss functions that primarily emphasize frame or pixel accuracy, often neglecting temporal coherency. DoP bridges this gap by focusing on pixel movement within videos (Figure \ref{fig:dop}), and optimizes 
the accuracy of pixel difference across frames with $(\mathcal{L}_{DoP})$,
\begin{equation}
\begin{aligned}
    \mathcal{L}_{DoP} &= \mathcal{L}_{u}^{DoP} + \mathcal{L}_{EoR}^{DoP} = MSE(\phi(t_{loc}), \phi(s_{loc})) \\ &+  MSE(\phi(t_{loc}^{EoR}), \phi(s_{loc})). 
    \label{eq:dop_loss}
\end{aligned}
\end{equation}
where, $\phi$ denotes temporal difference,
\begin{equation}
    \phi{(x^{f'})} = x^{f+1}_{loc} - x^{f}_{loc} 
    \label{eq:dop_cal}
\end{equation}
and, $x_{loc}^{f}$ means localization map at frame $f$.
This strategy enforces stronger \textit{temporal coherency} within spatio-temporal predictions and enhances the quality of pseudo-labels produced by the networks.

\noindent \textbf{Role of EoR and DoP:}  The base model provides a rough estimate of activity area.  EoR recognizes fine-grained errors in spatial boundaries and help as a enhanced class-agnostic supervision to improve student's ($\mathcal{M}_s$) spatio-temporal localization (Figure \ref{fig:brief_summary} left). However, EoR loss focuses on spatio-temporal localization without any temporal coherency which is important for action detection. This is where DoP plays a role and helps in temporal coherency of localization across frames. DoP constraint helps to generate a smooth flow of localization along temporal dimension as it enforces consistency on \textit{displacement of pixels}.

\noindent \textbf{Gradient flow:}
The base model and Error Recovery module are trained jointly but the gradients from the Error Recovery module are not used to update the base model. 
If the gradients are allowed to update the base model, then it will also impact the prediction of the base model (discussed in ablation study).  This will be same as adding more parameters to the model, which is not our goal. Our objective on the other hand is to learn from the mistakes of the base model and not to improve it. 
This also ensures that the improvement of pseudo-labels is not dependent on the input video and is class agnostic. The Error Recovery module will only have access to the prediction of the base model without any knowledge of the input video. This helps in learning a transformation that generalizes well to unlabeled samples.

\noindent \textbf{Augmentations:}
We study both spatial and temporal augmentations to generate weak and strong views. First, temporal and then spatial augmentations is applied on the input video. Augmenting in this sequence makes the process \textit{computationally efficient} as for spatial augmentation we only perform augmentation of required frames instead of augmenting all the video frames. The weak augmentation includes only horizontal flipping whereas strong augmentation includes color jitter, gaussian blur and grayscale.

\begin{table*}[t]
	\centering
	\renewcommand{\arraystretch}{1.06}
	\scalebox{1.0}{
		\begin{tabular}{r c | c | ccc | c | ccc}
			\specialrule{1.5pt}{0pt}{0pt}
			\rowcolor{mygray} 
   & & \multicolumn{4}{c|}{UCF101-24} & \multicolumn{4}{c} {JHMDB21}\\
    \rowcolor{mygray}  Method& Backbone  &  Annot. & f-mAP & \multicolumn{2}{c}{v-mAP}  & Annot. &  f-mAP & \multicolumn{2}{c}{v-mAP}   \\
    \hline
     \textbf{\textit{Fully-Supervised}} & & \% & 0.5  & 0.2 & 0.5 & \% & 0.5  & 0.2 & 0.5  \\
    \midrule 
      TACNet ~\cite{Song_2019_CVPR}$^{\dagger}$ & RN-50&& 72.1 & 77.5 & 52.9 & & 65.5 & 74.1 & 73.4  \\
       MOC ~\cite{li2020actions} & DLA-34 & &78.0 & 82.8 & 53.8 & &70.8 & 77.3 & 70.2\\
      ACAR-Net ~\cite{pan2021actor} & SF-R50&& 84.3 & -& - & & 77.9 & - & 80.1\\
     VideoCapsuleNet ~\cite{duarte2018videocapsulenet} & I3D& & 78.6 & \underline{97.1} & \underline{80.3}& & 64.6 & \underline{95.1} & -   \\
     YOWO \cite{yowo} & ResNext-101 & & 80.4 & 75.8 & 48.8 & &  74.4 & 85.7 & 58.1 \\
     TubeR ~\cite{Zhao_2022_CVPR}  &I3D& &83.2 & 83.3 & 58.4 &&  - & 87.4 & \underline{82.3}  \\
     STMixer ~\cite{Wu2023STMixerAO}  &  SF-R101NL & &83.7 & - & - && 86.7 & - & -\\

     EVAD~\cite{evad} &  ViT-B & & 85.1 & - & - && \underline{90.2} & - & -\\
     BMVIT ~\cite{bmvit} &  ViT-B & &\underline{90.7} & - & - && 88.4 & - & -\\
     
    \midrule
    \textbf{\textit{Weakly-Supervised}} &&&&& \\
    \midrule
     PSAL ~\cite{weakly5} & RN-50& &- &41.8 &- & &-&-&-\\
     Cheron \textit{et al.} \shortcite{cheron_2018} &RN-50  && - & 43.9  & 17.7 & &-&-&- \\
     GuessWA ~\cite{Escorcia2020GuessWA} & IRv2 && 45.8 & 19.3 & - &&-&-&- \\
     UAWS ~\cite{Arnab2020UncertaintyAwareWS} & RN-50  & &- & 61.7 & 35.0 & &-&-&- \\
     GLNet ~\cite{weakly4} & I3D & &30.4 & 45.5 & 17.3 & & 65.9 & 77.3 & 50.8 \\
    \hline
    \textit{\textbf{Semi-Supervised}} &&&&& \\
    \hline
    MixMatch ~\cite{mixmatch}$^{\dagger\dagger}$  & I3D & 10\% &10.3 & 54.7 & 4.9 & 30\% & 7.5 & 46.2 &  5.8\\
    Pseudo-label ~\cite{pseudocls} & I3D & 10\% &59.3 &89.9& 58.3 & 20\% & 55.3 & 87.6 & 52.0\\
    ISD ~\cite{co_ssd} & I3D& 10\% & 60.2 & 91.3 & 64.0 & 20\%  & 57.8& 90.2& 57.0 \\
    E2E-SSL  ~\cite{Kumar_2022_CVPR} & I3D& 10\% & 65.2 & 91.8  & 66.7 & 20\% & 59.1 & 93.2 & 58.7  \\
    \midrule
    Baseline Mean Teacher~\cite{mt} & I3D& 10\% & 67.3 & 92.7 & 70.5 & 20\% & 56.3 & 88.8 &52.8\\
    Stable Mean Teacher (Ours) & I3D &  10\% & \textbf{73.9} & \textbf{95.8} & \textbf{76.3} & 20\%  & \textbf{69.8} & \textbf{98.8} & \textbf{70.7}\\
    \hline
    \rowcolor{mygray} Supervised baseline & I3D & 10\% &53.5 & 77.2 & 49.7 & 20\%  & 55.7 & 93.9 & 52.4\\
			\specialrule{1.5pt}{0pt}{0pt}
	\end{tabular}}
 
 \caption{ \textbf{Comparison with previous state-of-the art approaches} on fully, weakly and semi-supervised learning on UCF101-24 and JHMDB21. $\dagger$ shows approach using Optical flow as second modality. The last row shows the score for supervised labeled subset, that is 10\% for UCF101-24 and 20\% for JHMDB21. Best score on each metric is underlined. RN-50, SF-R50/101 and IRv2 is ResNet-50, SlowFast-R50/101, and, InceptionResNetV2 respectively. \shortcite{mixmatch}$^{\dagger\dagger}$ suffers from cold-start problem below 30\% on JHMDB21.  }
	\label{tab:comparison_sota_ucf101}
\end{table*}

\subsubsection{Learning objectives}
\label{sec:optim}
The objective function of Stable Mean Teacher has two parts: supervised $(\mathcal{L}_{s})$ and unsupervised $(\mathcal{L}_{u})$. Supervised loss has classification $(\mathcal{L}_{s}^{cls})$ and localization $(\mathcal{L}_{s}^{loc})$ and follows losses from \shortcite{duarte2018videocapsulenet}. Unsupervised loss comprise of three parts: 1) Base model (STMT) loss $(\mathcal{L}_{base})$ which incorporates both classification $(\mathcal{L}_{base}^{cls})$ and localization $(\mathcal{L}_{base}^{loc})$, 2) Error Recovery loss $(\mathcal{L}_{EoR})$, and 3) DoP loss $(\mathcal{L}_{DoP})$. We calculate the supervised loss on the labeled subset of student's predictions $(\mathcal{L}_{s}^{cls}, \mathcal{L}_{s}^{loc})$ and student's Error Recovery module predictions, and unsupervised loss on labeled plus unlabeled subset. We have two unsupervised losses: \textbf{a) \textit{classification consistency}}: it minimizes the  difference between teachers' prediction $t_{cls}$ and student's prediction $s_{cls}$ using Jenson-Shennon Divergence (JSD), and, \textbf{b) \textit{localization consistency}}: computes pixel-level difference on each frame between teacher $(t_{loc}, t_{loc}^{EoR})$ and student $s_{loc}$ localization maps using MSE. Finally, the overall loss for Stable Mean Teacher is defined as,
\begin{equation}
    \mathcal{L} = \mathcal{L}_{s} + \lambda \mathcal{L}_{u} =  \mathcal{L}_{s} + \lambda (\mathcal{L}_{base} + \mathcal{L}_{EoR} + \mathcal{L}_{dop})
\end{equation}
 where $\lambda$ is a weight parameter for unsupervised losses.

\section{Experiments}
\noindent \textbf{Datasets:}  
We use four benchmark datasets to perform our experiments; UCF101-24 \shortcite{ucf101}, JHMDB21 \shortcite{jhmdb}, and AVA v2.2 (AVA)\shortcite{ava} for action detection, and YouTube-VOS \shortcite{vosdataset} to show generalization on video segmentation (VOS). UCF101-24 consists of 3207 videos, split into 2284 for training and 923 for testing. JHMDB21 has 900 videos with 600 for training and 300 for testing. The resolution of the video is 320x240 for both of the datasets. The number of classes in UCF101-24 is 24 and in JHMDB21 it's 21. AVA consists of 299 videos, each lasting 15 minutes. The dataset is divided into 211K clips for training and 57K clips for validation. Annotations are provided at 1 FPS with bounding boxes and labels. We report our performance on 60 action classes following standard evaluation protocols \cite{pan2021actor,zhao2021tuber}. The distribution of the number of training, validation and evaluation videos on YouTube-VOS-2019 \shortcite{xu2018youtube} is 3471,  507 and 541 respectively.

\noindent \textbf{Labeled and unlabeled setup:} 
The labeled and unlabeled subset for UCF101-24 and Youtube-VOS is divided in the ratio of 10:90 and for JHMDB21 it's 20:80. For AVA dataset, we use 50\% of the dataset for semi-sup setup. We utilize 10:40 split between labeled to unlabeled ratio. We perform our experiments with 10\%/20\% labeled set for UCF101-24/JHMDB21 instead of 20\%/30\% (as in \cite{Kumar_2022_CVPR}), as the performance is already close to fully supervised training when using 20\%/30\% in these datasets. These scores are shown in supplementary. 

\noindent \textbf{Implementation details} 
 We train the model for 50 epochs with a batch size of 8 where the number of samples from both labeled and unlabeled subsets are the same. The value of $\beta$ for EMA parameters update is set to 0.99 which follows prior works \shortcite{semiobj6, semiobj7}. The value of $\lambda$ for the unsupervised loss weight is set to 0.1 which is determined empirically. More details are provided in supplementary.
 
\begin{table*}[t!]
	\centering
	\renewcommand{\arraystretch}{1.06}
	\scalebox{1.0}{
		\begin{tabular}{c c c c c    c  c   c }
			\specialrule{1.5pt}{0pt}{0pt}
     \rowcolor{mygray} Method &  Backbone &  Pretraining & K & FPS & $\mathcal{A}$ & mAP & GFLOPs\\
    \multicolumn{8}{c }{\textit{Non real-time spatio-temporal action detector}}   \\
    \hline
    WOO \shortcite{woo}  & SF-R101 & K600& 8 & - & 100\% & 28.3 & 252\\
     SE-STAD \shortcite{sestad} & SF-R101 & K400 & 8 & - &100\% &  29.3 & 165\\
     TubeR \shortcite{zhao2021tuber} & CSN-152 & IG-65M &32 & 3 &100\% &29.7 & 120\\
     STMixer \shortcite{Wu2023STMixerAO} & CSN-152 & IG-65M & 32 & 3 & 100\% &31.7 & 120 \\
     EVAD \shortcite{evad} & ViT-B & K400 & 16 & - &100\% & 32.3 & 243 \\
     BMViT \shortcite{bmvit} & ViT-B & K400, MAE & 16 & - & 100\% &31.4 & 350 \\
     \hline
     \multicolumn{8}{c }{\textit{Real-time spatio-temporal action detector}}   \\
     \hline
     YOWO \shortcite{yowo} & ResNext-101 &  K400 & 16 & 35 & 100\% &17.9 & 44\\
     YOWOv2-N \shortcite{yowov2} & Shufflev2-1.0x & K400 & 16 & 40 & 100\% &12.6 & 1.3 \\
     \hline
     Ours(YOWOv2-N) & Shufflev2-1.0x & K400 & 16 & 40 & 10\% &8.5 & 1.3 \\
     \hline
     \rowcolor{mygray} Sup. baseline & Shufflev2-1.0x & K400 & 16 & 40 & 10\% &5.2 & 1.3\\
    \specialrule{1.5pt}{0pt}{0pt}
	\end{tabular}}
 
 \caption{ \textbf{Evaluation on AVA dataset.} K is the length of input video clip. $\mathcal{A}$ denotes annotation percent. mAP denotes f-mAP@0.5. YOWO2-N denotes nano version.
  }
  \label{tab:ava_results}
\end{table*}

\noindent \textbf{Base model and Error Recovery model architecture:} 
We use VideoCapsuleNet \shortcite{duarte2018videocapsulenet} as our base action detection model. It is a simple encoder-decoder based architecture that utilizes capsule routing. Different from the original model, we use 2D routing instead of 3D routing which makes it computationally efficient. This also maintains consistency with the previous work \shortcite{Kumar_2022_CVPR} and enables a fair comparison. For the Error Recovery module, we use a 3D UNet \shortcite{unet3d} architecture with a depth of 4 layers with 16, 32, 64 and 128 channels respectively. 

%
 
\noindent \textbf{Evaluation metrics:}
For spatio-temporal video localization, we evaluate the proposed approach similar to previous works \shortcite{finn2016unsupervised, weinzaepfel2015learning} on frame metric average precision (f-mAP) and video metric average precision (v-mAP). f-mAP is computed by summing over all the frames with an IoU greater than a threshold per class. Similarly, for v-mAP 3D IoU is utilized instead of frame-level IoU. We show results at 0.2 and 0.5 thresholds in the main paper with other thresholds results provided in the supplementary. For VOS, we show results on Jaccard $(J)$ and Boundary $(F)$ metrics.

\begin{table}[t]
	\centering
	\renewcommand{\arraystretch}{1.06}
	\scalebox{1.0}{
		\begin{tabular}{c|c|c| c |  c | c | c }
			\specialrule{1.5pt}{0pt}{0pt}
    \rowcolor{mygray} & & & \multicolumn{2}{c| }{UCF101-24} & \multicolumn{2}{c }{JHMDB-21} \\
     \rowcolor{mygray} $\mathcal{L}_{base}$ & $\mathcal{L}_{EoR}$ & $\mathcal{L}_{DoP}$ &v@0.5 & f@0.5 &v@0.5 & f@0.5\\
    \hline
      \checkmark& & & 74.5 & 72.2 & 62.0 & 61.8  \\
       \checkmark&\checkmark &    &  75.9 & 73.1 &  68.1 & 68.3   \\
      \checkmark& & \checkmark   & 75.4 & 72.6  & 64.5 & 62.9  \\
    \checkmark & \checkmark& \checkmark &  76.3  & 73.9  & 70.7 & 69.8  \\
    \specialrule{1.5pt}{0pt}{0pt}
	\end{tabular}}
 
 \caption{ \textbf{Ablations:} Effectiveness of \textit{Error Recovery module} and \textit{Difference of Pixels}. $\mathcal{L}_{base}$: training without $\mathcal{L}_{EoR}$ and $\mathcal{L}_{DoP}$. v@0.5:  v-mAP@0.5, f@0.5:f-mAP@0.5.
  }
  \label{tab:ablation_aux}
 
\end{table}

\subsection{Results}
\noindent \textbf{Comparison with semi-supervised:} In Table \ref{tab:comparison_sota_ucf101}, amongst \textit{semi-supervised} approaches, the first two rows show the performance of image-based strategy, while the third row shows the performance by an object detection approach, and lastly, \shortcite{Kumar_2022_CVPR} is a pi-consistency based technique for video action detection. \shortcite{mixmatch} is not able to generalize well with less amount of videos. 
Our proposed approach beats the pseudo-label based approach on all thresholds. Compared with the semi-supervised object detection approach, we outperform it by 12-14\% on UCF101-24 and 9-12\% on JHMDB21 using 10\% less data.  Furthermore, comparing with a parallel approach to semi-supervised video action detection, we have a gain of 8.7\% on f-mAP@0.5 and 9.6\% on v-mAP@0.5 on the UCF101-24 dataset. On JHMDB21, the gain is 5.4\% and 7.2\% at f-mAP@0.5 and v-mAP@0.5 respectively with 10\% less data. Against our base model without DoP and EoR modules, our proposed approach have an improvement of 1.7, 0.7, and, 1.8 on UCF101-24, and, 7.0, 4.1, and, 8.7 at f-mAP@0.5, v-mAP@0.2 and v-mAP@0.5 respectively. 

\noindent \textbf{Comparison with supervised and weakly-supervised:} We start with \textit{supervised} scenario where with only 10\% labeled data, our performance surpasses all the 2D-based approaches on v-mAP (Table \ref{tab:comparison_sota_ucf101}). Amongst 3D-based methods, we outperform a few of them and show competitive performance with others. Worth noting is that while most 2D approaches incorporate optical flow as a secondary modality, our architecture stands out by relying solely on a single modality. Shifting focus to \textit{weakly-supervised}, our method surpasses the state-of-the-art on both datasets by a substantial margin. Comparing against the best approach \cite{Arnab2020UncertaintyAwareWS, weakly4}, on UCF101-24 (Table \ref{tab:comparison_sota_ucf101}), our approach outperforms by an approximate margin of approx. 35\% at the 0.5 thresholds. On JHMDB21 (Table \ref{tab:comparison_sota_ucf101}), we observe a significant enhancement, with an absolute boost of 3.9\% on f-mAP@0.5 and 19.9\% on v-mAP@0.5. 

\noindent \textbf{Scaling to large-scale dataset:} To evaluate the scalability of our approach, we perform experiments on AVA dataset a large scale dataset. AVA is not spatio-temporal as against UCF101-24 with only sparse frame level annotations available. In Table \ref{tab:ava_results}, we run on a real-time spatio-temporal approach and show our approach improves on supervised by 3.3\% on YOWOv2-N with only 10\% labeled data.

\begin{figure}[t!]
\centering
\includegraphics[width=\linewidth]{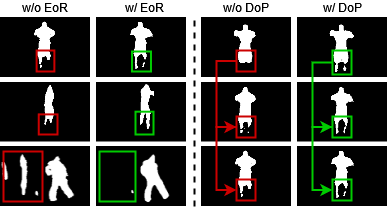}

\caption{ \textbf{Qualitative analysis for EoR and DoP:}  Left side illustrates the effectiveness of \textit{Error Recovery} module on multiple samples, with improvement in action boundary precision and it also helps in suppressing background noise.  On the right hand, we demonstrate how \textit{DoP constraint} induces temporal coherency in predictions for sequence of video frames. 
}
 
\label{fig:brief_summary}
\end{figure}

\begin{figure*}[t!]
    \centering
    \includegraphics[height=0.13\textheight]{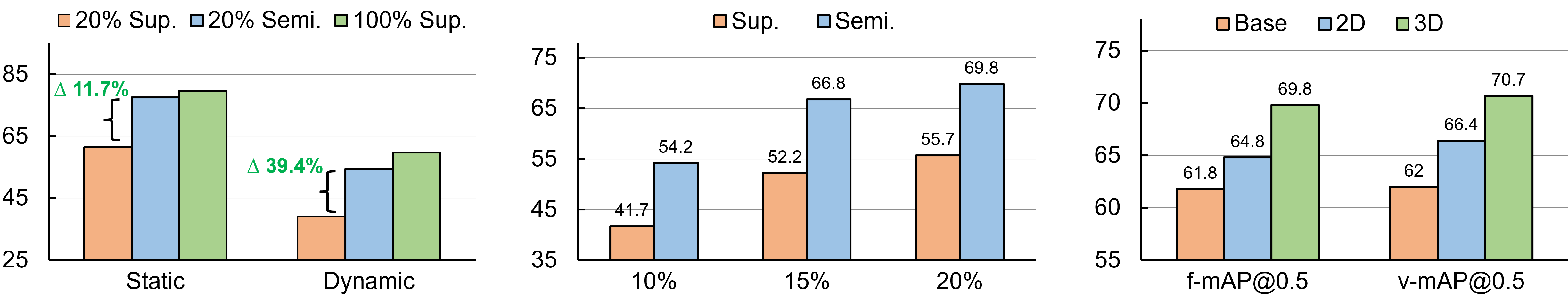}
    \caption{ \textbf{\textit{Analyzing Stable Mean Teacher:}} \textit{(Left)} \textit{\textbf{Static vs dynamic scenes:}}
    Dynamic scenes are challenging than static scenes, however, the relative boost in performance for dynamic is 27.7\% more than in case of static scene scenario. $\Delta$ denotes relative change at v-mAP@0.5. \textit{(Middle)} \textbf{\textit{Annotation percent:}} Moving towards right to left on x-axis, the gain in performance (f-mAP@0.5) increases. It indicates the approach is more effective in low label regime. \textit{(Right)} \textit{\textbf{Error Recovery architectures:}} The performance of 3D Error Recovery architecture outperforms the 2D based architecture.
    }  
     
    \label{fig:discussion}
\end{figure*}

\begin{figure}[t!]
\centering
        \includegraphics[width=\linewidth]{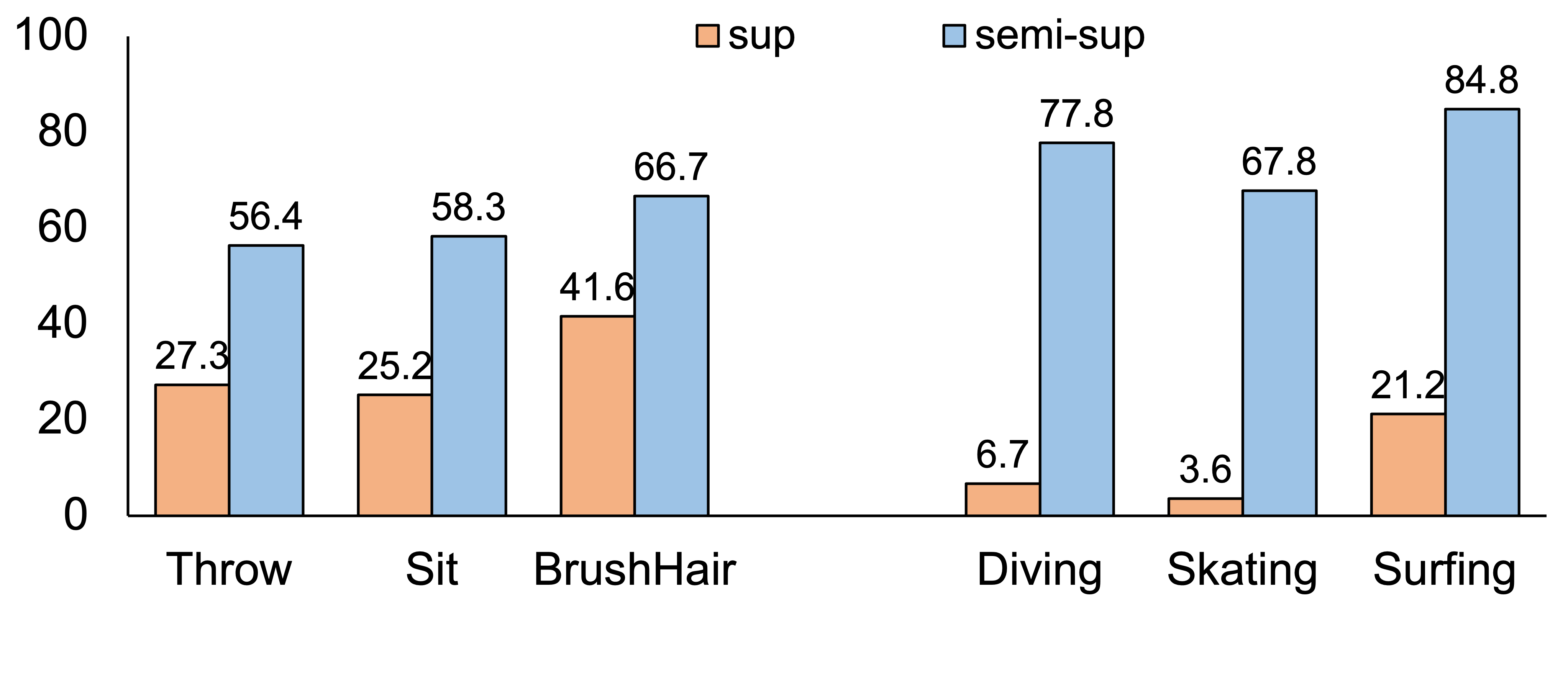}
 \caption{ \textbf{Classwise analysis:}  Improvement in v-mAP@0.5 for top 3 action classes with maximum performance gain over supervised baseline on static: \textit{\{\texttt{throw, sit, brushhair}\}} and dynamic \textit{\{\texttt{diving, skating, surfing}\}} showing effectiveness of the proposed approach. 
}
  
\label{fig:top_classes}
\end{figure}

\subsection{Ablation studies}

\noindent \textbf{\textit{Impact of Error Recovery module:}} We begin by highlighting the significance of the EoR module in Table \ref{tab:ablation_aux}. A comparison between the second and first rows reveals a substantial performance boost of 6\% and 1\% over the baseline architecture on JHMDB-21 and UCF101-24 respectively. This enhancement is attributed to the refined pseudo labels $(t_{loc}^{EoR})$, which provide superior guidance to the student $(s_{loc})$ in terms of more precisely \textit{localized} activity regions. We observe a major boost for JHMDB-21 against UCF101-24 since it's a more challenging dataset with pixel-level ground truth as against bounding box level ground truth in UCF101-24. We extended this analysis to a higher average v-map@0.5:0.95 to delve into its impact on a finer level. The results show an improvement of 2.5\% and 3.9\% in mean IoU for f-mAP@0.5:0.95 and v-mAP@0.5:0.95, respectively. This underscores the proposed approach's ability to enhance finer boundary regions.

\noindent \textbf{\textit{Effect of DoP constraint:}} The incorporation of the Difference of Pixels (DoP) constraint is aimed at introducing temporally coherent pseudo labels. As evident from Table \ref{tab:ablation_aux}, it provides improvements both in conjunction with the base model (STMT) and when used alongside the Error Recovery module, each by a margin of 0.5-0.8\% and 1-3\% on UCF101-24 and JHMDB-21 respectively. Notably, the proposed DoP constraint exhibits a more pronounced enhancement in v-map compared to f-map, indicating its positive influence on \textit{temporal coherency} within predictions. When employed on its own, the DoP constraint also results in a 1\% increase in mean IoU for f-mAP@0.5:0.95 and a 2\% increase in v-mAP@0.5:0.95, further underlining its efficacy. 

\noindent \textbf{\textit{Qualitative analysis:}} 
In Figure \ref{fig:brief_summary}, we analyze the effectiveness of each component qualitatively. DoP makes the predictions coherent across time and Error Recovery module helps to generate better fine-grained predictions. We show more qualitative analysis in supplementary.

\subsection{Discussion and analysis} 
\label{sec:discussion}
We further answer some of the important set of questions pertaining to Stable Mean Teacher approach for semi-supervised activity detection in this section.

\noindent \textbf{\textit{Static vs dynamic scenes:} }
Activities can be categorized into two sub-classes based on the background namely static, where background is constant, and dynamic, when the background changes. To give an overview, example of few classes categorized as static and dynamic, in JHMDB21, are \textit{\{\texttt{brush\_hair, golf, pour, shoot\_bow, sit}\}} and \textit{\{\texttt{climb\_stairs, jump, run, walk, push}\}} respectively. Dynamic is a challenging situation since the actor and background both is changing in each frame. Our proposed solution makes a \textbf{11.7\%} improvement on static and \textbf{39.4\%} on dynamic actions at v-mAP@0.5 (Figure \ref{fig:discussion} (left)). 
This demonstrates the effectiveness of our approach for dynamic videos. We show classwise analysis in Fig.\ref{fig:top_classes}.

\noindent \textbf{\textit{Effectiveness of approach in low-label regime:}} In Figure \ref{fig:discussion} (middle), we look into performance gain at multiple ratios of labeled and unlabeled. Comparing between  15\% vs 20\%, the gain at 15\% is 40\% more than the gain at 20\% which shows that our approach is even more effective in low-label data regime and it uses unlabeled data more effectively.

\noindent \textbf{\textit{Error Recovery architectures:}} We analyze the effect of different architectures on Error Recovery module. With a 2D CNN backbone we observe that the performance degrade by an absolute margin of 3\% at f-mAP@0.5 and 4\% at v-mAP@0.5 (shown in Figure \ref{fig:discussion} (right)) which supports 3D CNN which generates better spatio-temporal pseudo labels.

\noindent \textbf{\textit{Importance of gradient stopping:}} Error Recovery module utilize grayscale maps to localize the actor whereas main model uses RGB frames to classify and localize the action. Since the task of Error Recovery module is to be class-agnostic, if gradient of Error Recovery module flows back into  the main network, it degrades the quality of pseudo labels generated by the main model. This further degrades the refinement procedure by Error Recovery module. We observe performance degrades by approximately 3\% without stopping the gradient flow.

\noindent \textbf{\textit{Additional parameters doesn't help:}} In this study, we add the EoR module parameters to the base model. The performance on JHMDB-21 at 20\% for f-mAP@0.5 is 62.3 and at v-mAP@0.5 it's 63.4. The model shows some improvement over the base model (Table \ref{tab:comparison_sota_ucf101}) by a margin of 0.5-1\% due to additional parameters. However, comparing it to proposed approach, it still lacks by \~7\%. This shows that simple extension by adding additional parameter doesn't help as such.

\begin{table}[t!]
   \centering
   \renewcommand{\arraystretch}{1.06}
  \scalebox{1.0}{
  \begin{tabular}{ c | c |c |c c c c}
    \specialrule{1.5pt}{0pt}{0pt}
    \rowcolor{mygray} Method& Annot. & Avg & $J_{S}$ & $J_{U}$ & $F_{S}$ & $F_{U}$\\
    \hline
    Xu  \shortcite{voslstm} $^{\dagger}$ & 10\% &  10.1  &  11.6  &  10.1  &  9.6  & 9.2 \\
    Kumar \textit{et al.} \shortcite{Kumar_2022_CVPR}& 10\% &  36.8 & 43.1 & 31.4   &  40.8 & 31.8 \\
    \textbf{Ours} & 5\% & 38.2  & 45.3 & 32.0 &   43.2  &  32.2\\
     \textbf{Ours} & 10\% & 41.3  & 48.2 & 35.0 &   46.7  &  35.4 \\
     \hline
    Xu  \shortcite{voslstm} & 100\% & 47.9 & 55.7 & 39.6 & 55.2 & 41.3 \\
    \specialrule{1.5pt}{0pt}{0pt}
  \end{tabular}}
  
  \caption{  \textbf{Generalization capability:} Performance comparison on Youtube-VOS. $J_{s}$ and $J_{u}$ are Jaccard on seen and unseen categories; $F_{s}$ and $F_{u}$ are boundary metric on seen and unseen categories. $^{\dagger}$ shows 10\% supervised results.}
  \label{tab:youtube_vos}
  
\end{table}

\paragraph{Generalization to video object segmentation (VOS)}
We further demonstrate the generalization capability of Stable Mean Teacher on VOS. We perform our experiments on YouTube-VOS dataset for this experiment and the results are shown in Table \ref{tab:youtube_vos}. We observe that the proposed method outperforms the supervised baseline by an absolute margin of 31\% on average against labeled setup. Comparing to the semi-supervised approach \shortcite{Kumar_2022_CVPR}, our approach shows a gain of 4-6\% on all metrics. Even with half of labeled data, at 5\% labeled data, our proposed approach beats \shortcite{Kumar_2022_CVPR}.
\section{Conclusion}

We propose \textit{Stable Mean Teacher}, a novel student-teacher approach for semi-supervised action detection. 
Stable Mean Teacher relies on a novel \textit{Error Recovery} module which learns from  student's mistakes and transfer that knowledge to the teacher for generating better pseudo labels for the student. It also benefits from \textit{Difference of Pixels}, a simple constraint which enforces temporal coherency in the spatio-temporal predictions. 
We demonstrate the effectiveness of Stable Mean Teacher on three action detection datasets with extensive set of experiments. Furthermore, we also show its performance on VOS task validating its \textit{generalization capability} to other dense prediction tasks in videos.

\newpage


\section{Stable Mean Teacher for Semi-supervised Video Action Detection \\ (Supplementary)}

Here, we go through some additional quantitative and qualitative results, extra ablation studies and, implementation details. Section I provides quantitative results on baseline setup. Section  II provides comparison at different percentages against previous semi-supervised approaches. Section III provides some more discussions on JHMDB21 and UCF101-24. Section IV discuses some implementation details mentioned in main paper. Section V shows extra qualitative results.

\section{Baseline Setup}
\label{sec:baseline}
We setup the baseline spatio-temporal mean teacher (STMT) and show it's comparison to baseline Mean teacher and our proposed approach. We observe an improvement of approx. 5\% f-mAP@0.5 on UCF101-24 and 4.5\% on JHMDB-21. At v-mAP@0.5, the gain 4-5\% on both the datasets. The final proposed approach further boost the score on top of baseline STMT model by a margin of 1.6\% on UCF101-24 and 7\% on JHMDB-21 at f-mAP@0.5. At v-mAP@0.5, the gain is 1.8\% on UCF101-24 and 8\% on JHMDB-21.

\begin{table*}[t]
	\centering
	\renewcommand{\arraystretch}{1.06}
	\scalebox{1.0}{
		\begin{tabular}{r | c | c | cc | c | c | ccc}
			\specialrule{1.5pt}{0pt}{0pt}
			\rowcolor{mygray} 
   & \multicolumn{4}{c|}{UCF101-24} & \multicolumn{4}{c} {JHMDB21}\\
   \rowcolor{mygray} Method  &   Annot. & f-mAP & \multicolumn{2}{c|}{v-mAP}  & Annot. &  f-mAP & \multicolumn{2}{c}{v-mAP}   \\
    & \% & 0.5  & 0.2 & 0.5 & \% & 0.5  & 0.2 & 0.5  \\
    \hline
    Baseline Mean Teacher\cite{mt} & 10\% & 67.3 & 92.7 & 70.5 & 20\% & 56.3 & 88.8 &52.8\\
    ST-MeanTeacher (Ours) & 10\% & 72.2 &95.1 & 74.5 & 20\%  & 61.8 & 94.7& 62.0 \\
    Stable Mean Teacher (Ours) &  10\% & \textbf{73.9} & \textbf{95.8} & \textbf{76.3} & 20\%  & \textbf{69.8} & \textbf{98.8} & \textbf{70.7}\\
    \hline
    Supervised baseline & 10\% &53.5 & 77.2 & 49.7 & 20\%  & 55.7 & 93.9 & 52.4\\
			\specialrule{1.5pt}{0pt}{0pt}
	\end{tabular}}
 \caption{ Analysis on Baseline Spatio-temporal mean teacher (STMT). STMT vs Stable Mean Teacher.}
  \label{tab:comparison_stmt_smt}
\end{table*}

\section{Comparison with previous Semi-supervised approaches }
\label{sec:sota}

We extended the Table 1 in main paper to have detailed comparison on more semi-supervised approaches at multiple labeled percentages on UCF101-24 and JHMDB21 in Tables \ref{tab:comparison_sota_ucf101_multi} and \ref{tab:comparison_sota_jhmdb21_multi} respectively. 

We show that our proposed approach outperforms previous semi-supervised approaches on multiple thresholds. 

\begin{table*}[t]
	\centering
	\renewcommand{\arraystretch}{1.06}
	\scalebox{1.0}{
		\begin{tabular}{r | cc| cc | cc| cc}
			\specialrule{1.5pt}{0pt}{0pt}
			\rowcolor{mygray} 
   & \multicolumn{2}{c}{5\%} & \multicolumn{2}{c} {8\%} & \multicolumn{2}{c} {10\%} & \multicolumn{2}{c} {15\%}\\
   \rowcolor{mygray} Method  &   Annot. & f-mAP & \multicolumn{2}{c|}{v-mAP}  & Annot. &  f-mAP & \multicolumn{2}{c}{v-mAP}   \\
   \rowcolor{mygray} & f@0.5 &v@0.5& f@0.5 &v@0.5& f@0.5 &v@0.5& f@0.5 &v@0.5\\
    \hline
    Le \textit{et al.} \cite{pseudocls} &52.5 & - & 56.4 & - &59.3 & 58.3&64.9 &-\\
    Jeong \textit{et al.} \cite{co_ssd}& 53.2 & 51.5 & 57.8& 61.2 & 60.2 & 64.0& 63.9&65.2\\
    Kumar \textit{et al.}  \cite{Kumar_2022_CVPR}  & 59.1 & 58.8  & 64.1 & 64.6 & 65.2 & 66.7 & 68.3 & 70.2  \\
    \hline
     Tarvainen \textit{et al.} \cite{mt} & 58.4 & 59.2 &66.2&69.3 & 67.3 & 70.5 & 71.5 &74.7\\
    Ours & 66.2 & 68.2 & 72.1 & 73.8 & 73.9 & 76.3 & 75.5 & 77.9\\
    \hline
    Supervised baseline & 37.5 & 30.1 & 42.6 & 39.4 &53.5  & 49.7  & 58.8 & 55.6\\
			\specialrule{1.5pt}{0pt}{0pt}
	\end{tabular}}
 \caption{ \textbf{Comparison with previous state-of-the art approaches} on semi-supervised learning on UCF101-24 on multiple labeled subset.}
  \label{tab:comparison_sota_ucf101_multi}
\end{table*}

\begin{table*}[t]
	\centering
	\renewcommand{\arraystretch}{1.06}
	\scalebox{1.0}{
		\begin{tabular}{r | cc| cc | cc| cc}
			\specialrule{1.5pt}{0pt}{0pt}
			\rowcolor{mygray}  & \multicolumn{2}{c}{10\%} & \multicolumn{2}{c} {15\%} & \multicolumn{2}{c} {20\%} & \multicolumn{2}{c} {25\%}\\
    \rowcolor{mygray} & f@0.5 &v@0.5& f@0.5 &v@0.5& f@0.5 &v@0.5& f@0.5 &v@0.5\\
    \hline
    Jeong \textit{et al.} \cite{co_ssd}& 48.5 & 46.2 &54.3 & 51.8 & 57.8 & 57.0& 59.5 & 58.1\\
    Kumar \textit{et al.}  \cite{Kumar_2022_CVPR}  & 45.8 & 41.8 &52.1 & 50.1&59.1 & 58.7 & 61.4 & 60.5 \\
    \hline
     Tarvainen \textit{et al.} \cite{mt} & 47.6 & 44.3 & 60.9& 60.4& 61.8 & 62.0 & 65.4 & 66.0\\
    Ours & 54.2 & 50.3  & 66.8 & 66.5   & 69.8 & 70.7  &71.3 & 70.9\\
    \hline
    Supervised baseline & 43.7& 37.7 &50.2 &48.6  & 55.7 & 52.4 & 60.4 & 59.3\\
			\specialrule{1.5pt}{0pt}{0pt}
	\end{tabular}}
 \caption{ \textbf{Comparison with previous state-of-the art approaches} on semi-supervised learning on JHMDB21 on multiple labeled subset.}
  \label{tab:comparison_sota_jhmdb21_multi}
\end{table*}

\section{Extra Discussions}
\label{sec:ablation}

Firstly, we extend the same discussions as mentioned in main paper to UCF101-24 dataset on static vs dynamic, network architecture and performance improvement in low-label regime.

\paragraph{\textit{Static vs Dynamic scenes}} 
Activities can be categorized into two sub-classes based on the background namely  static, where background is constant, and dynamic, when the background changes. To give an overview, example of few classes categorized as static and dynamic, in UCF101-24, are \texttt{basketball, golf swing, rope climbing, soccer juggling, tennis swing} and \texttt{basketball dunk, biking, cliff diving, diving, skateboarding} respectively. Dynamic is a challenging situation since the actor and background both is changing in each frame. It's also evident from lower results as compared to Static actions from Table \ref{tab:abla_bkgrnd} on supervised settings. Our proposed solution makes a \textbf{18.8\%} improvement on static and \textbf{23.3\%} on dynamic actions at f-mAP@0.5. This shows that proposed approach is able to localize actor in challenging situations as well.  

\paragraph{\textit{Analysis on ErrOr Recovery (EoR) Module}} Here, we analyze the effect of different architectures of EoR network. We replace the EoR network with a 2D version and saw that the performance degrade by a small margin (Table \ref{tab:abla_arch}) which supports 3D based version generates better spatio-temporal pseudo labels similar to results shown on JHMDB21. 

\paragraph{\textit{Effectiveness of approach in low-label regime}} Similar to JHMDB21, we look into performance gain at multiple ratios of labeled and unlabeled on UCF101-24. Looking into Table \ref{tab:abla_label}, we see that we have huge performance gain much more evident than JHMDB21 at lower percentages. The gain at 5\% and 8\% is almost double the gain at 10\%.

\paragraph{\textit{Burn-in vs end-to-end?}} 
Some recent approaches have shown the benefit of pre-training on labeled set for model initialization \cite{semiobj7}. 
In this experiment, we analyzed the effect of burn-in on proposed Stable Mean Teacher and pre-trained the model on labeled set and use it for combined training on labeled and unlabeled sets.  
We observe that there was \textit{no substantial gain} to Stable Mean Teacher with burn-in weights.

\paragraph{\textit{Comparison on more thresholds}} Here, we extend ablation table on EoR Module and compare the \textit{improvement by sub modules} on JHMDB21 dataset. We show results on three more thresholds at 0.3, 0.4 and 0.6. Table \ref{tab:ab_aux_more_thresh} shows that proposed sub-modules helps in improvement more for higher thresholds. 

 \begin{table*}[t]
 \centering
    \captionsetup[subtable]{font=normalsize}

	\begin{subtable}[t]{0.25\textwidth}
		\centering
		\tablestyle{2pt}{1.02}
		\small
  \caption{\textbf{Static vs Dynamic.} 
		\begin{tabular}{c|cc}
\rowcolor{mygray} Category & Static & Dynamic   \\ \shline
20\% sup.  &  67.5 & 36.9  \\
20\% semi  & 86.3 & 60.2    \\
100\% sup. & 87.4 & 65.1    
\end{tabular}}
		\label{tab:abla_bkgrnd}
	\end{subtable}
        \begin{subtable}[t]{0.4\textwidth}
		\tablestyle{2pt}{1.02}
		\small
  \caption{\textbf{Network Architecture.}}
		\begin{tabular}{c|cc}
   \rowcolor{mygray}  Arch.  & f@0.5 & v@0.5 \\ \shline
     Base & 73.4 & 75.8 \\
2D   & 73.6 & 76.0 \\
3D & 73.9 & 76.3
\end{tabular}
	\label{tab:abla_arch}
	\end{subtable}
	\begin{subtable}[t]{0.25\textwidth}
		\tablestyle{2pt}{1.02}
		\small
  	\caption{\textbf{Low label regime}}
		\begin{tabular}{l|ccc}
   \rowcolor{mygray} & Sup. & Semi. & $\uparrow$ \%age          \\ \shline
    5\%  & 37.5 & 66.2 & 76.5\%\\
    8\%  & 42.6 & 72.1 & 69.2\% \\
    10\%  & 53.5 & 73.9 & 38.1\% \\
    \end{tabular}
		\label{tab:abla_label}
	\end{subtable}
\caption{\textbf{Analysis on UCF101-24} on multiple factors. f@0.5 and v@0.5 denotes f-mAP@0.5 and v-mAP@0.5 respectively. Tables \ref{tab:abla_bkgrnd} and \ref{tab:abla_label} shows performance on f-mAP@0.5.
}
	\label{tab:ablations}
\end{table*}

\begin{table*}[t]
	\centering
	\renewcommand{\arraystretch}{1.1}
	\scalebox{1.1}{
		\begin{tabular}{ ccc | cccc  | cccc }
			\specialrule{1.5pt}{0pt}{0pt}
			\rowcolor{mygray} L2& EoR & DoP &\multicolumn{4}{c|}{f-mAP} & \multicolumn{4}{c}{v-mAP}\\
   \hline
    &  &  & @0.3 & @0.4& @0.5 & @0.6  & @0.3 & @0.4& @0.5 & @0.6  \\
    \hline
     \checkmark& & & 83.6 & 74.2 &  61.8 & 46.4 & 92.8 & 79.7 & 62.0 & 38.2\\
      \checkmark & \checkmark &  & 89.2 & 81.2 & 68.3 & 49.6 &96.8 & 85.2 & 68.1 & 42.8\\
      \checkmark &  & \checkmark & 84.4 & 75.0 & 62.9 & 46.5 &90.7 &82.2& 64.5&39.3\\
    \checkmark & \checkmark & \checkmark &91.9 & 84.6 & 69.8 & 50.5 & 95.9 &85.6&70.7&42.9\\
			\specialrule{1.5pt}{0pt}{0pt}
	\end{tabular}}
\caption{\textbf{Ablation study on sub-modules at multiple thresholds}. L2: Base mean teacher model with L2 as spatio-temporal localization loss, EoR: EoR network, and DoP: Difference of pixel constraint.  }
  \label{tab:ab_aux_more_thresh}
\end{table*}

 \begin{figure*}[t!]
    \centering
    \begin{subfigure}{0.48\linewidth}
       \centering
      \includegraphics[width=\linewidth]{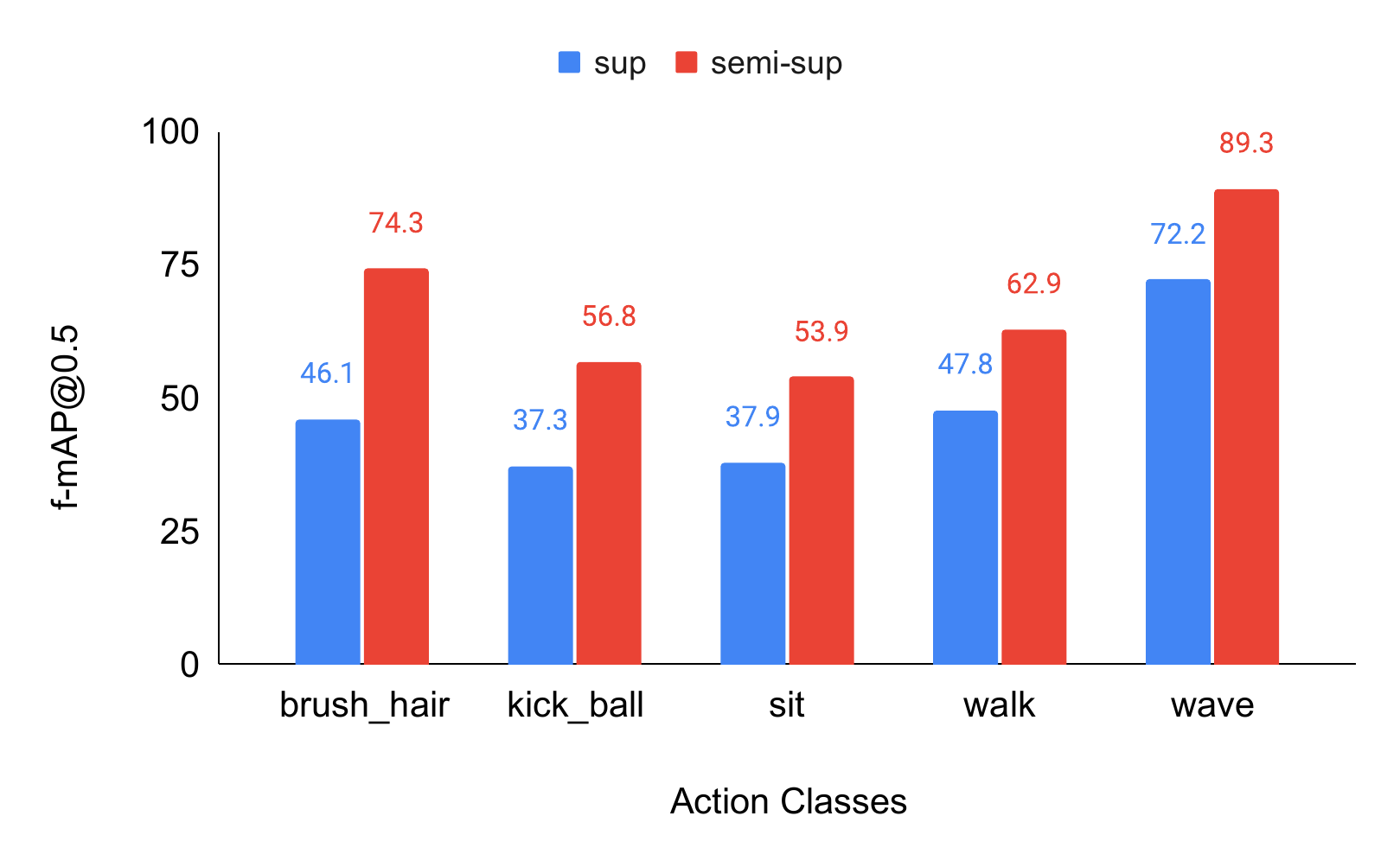}
    \end{subfigure}
    \begin{subfigure}{0.48\linewidth}
       \centering
      \includegraphics[width=\linewidth]{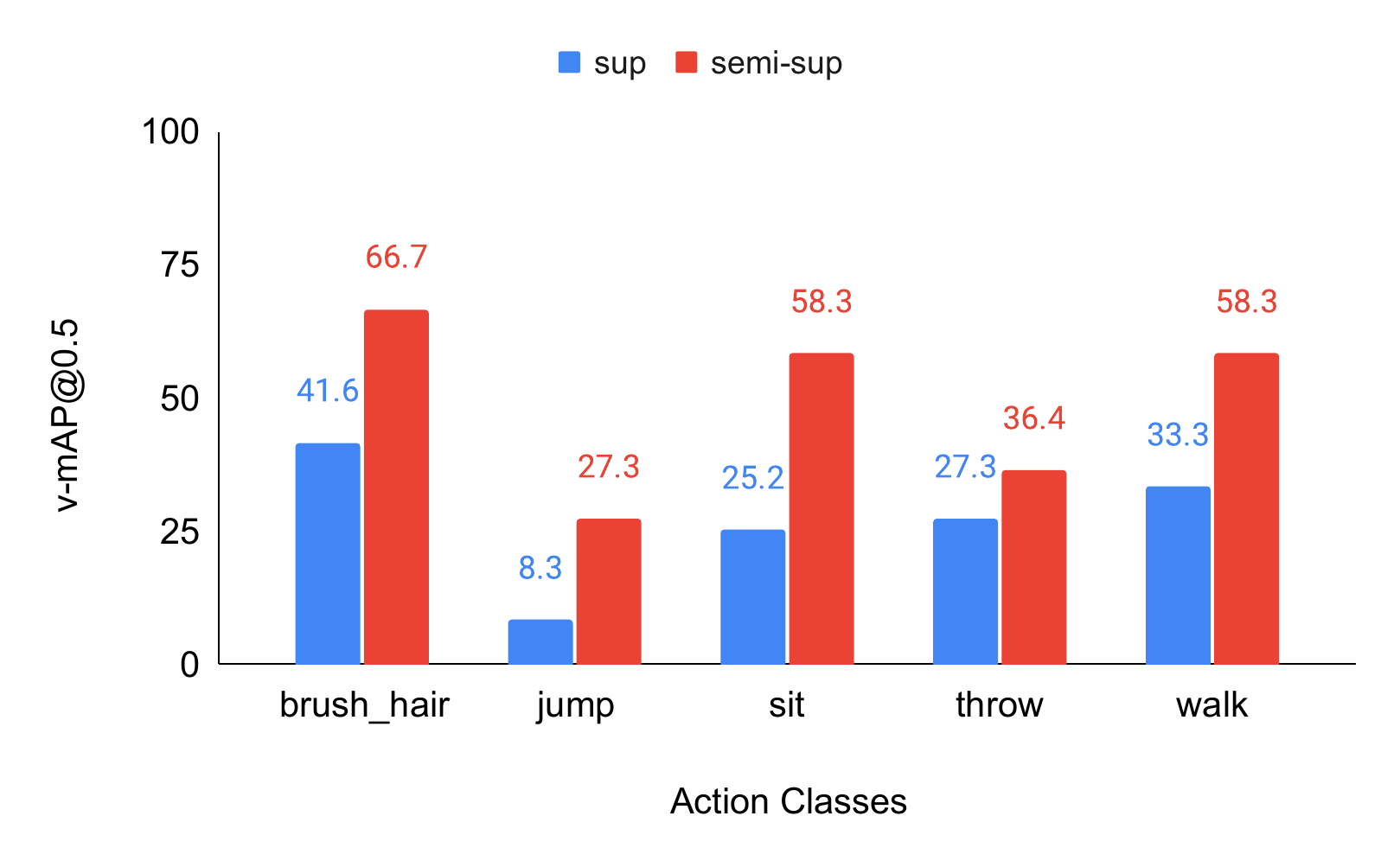}
    \end{subfigure}
    \caption{This figure shows the top 5 classes which has the most improvement on v-mAP@0.5 on our proposed semi-supervised approach compared to the supervised counterpart on JHMDB21 dataset. }
    \label{fig:jhmdb_vmap_fmap_best}
\end{figure*}

\begin{figure*}[t!]
    \centering
    \begin{subfigure}{0.48\linewidth}
       \centering
      \includegraphics[width=\linewidth]{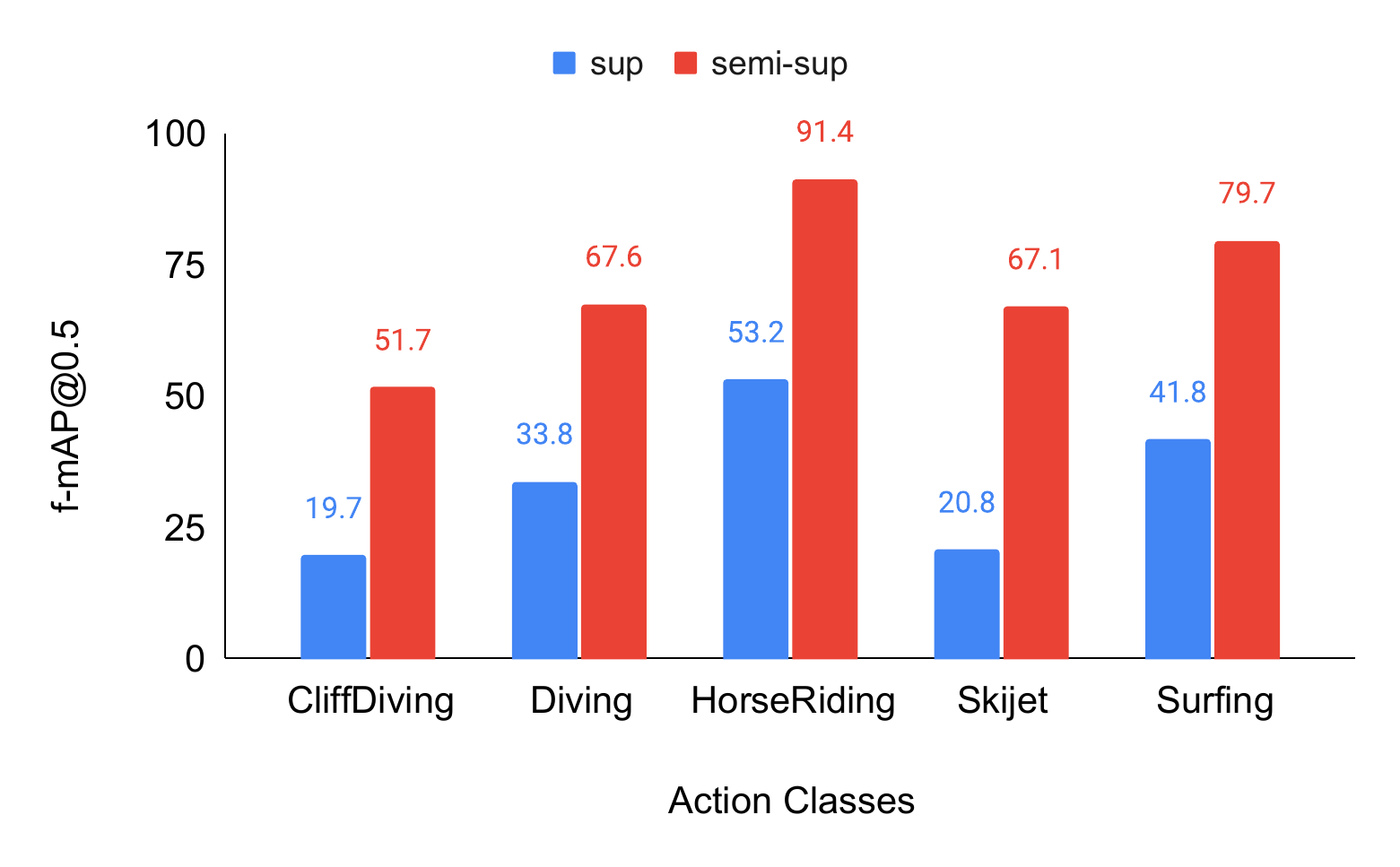}
    \end{subfigure}
    \begin{subfigure}{0.48\linewidth}
       \centering
      \includegraphics[width=\linewidth]{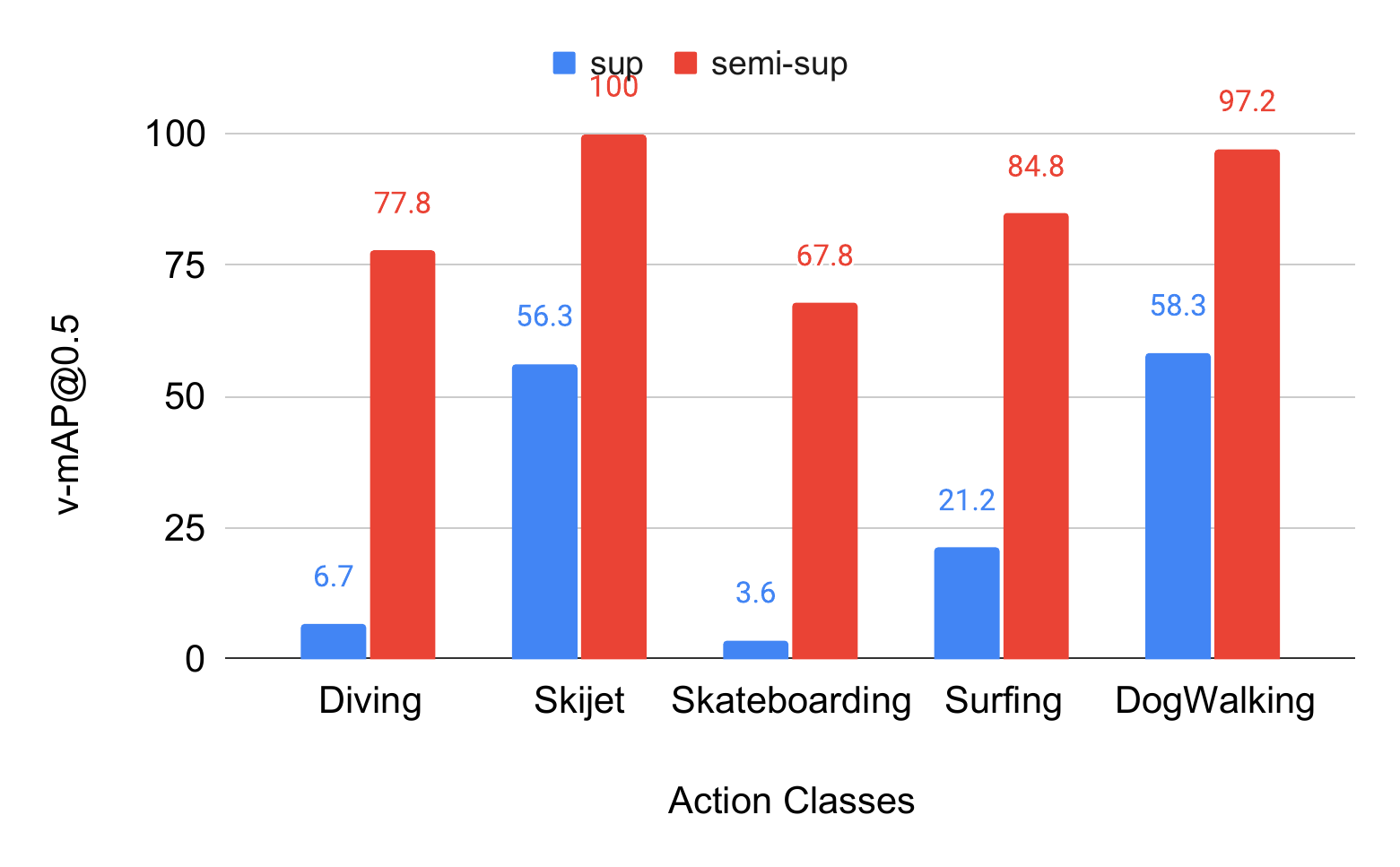}
    \end{subfigure}
    \caption{This figure shows the top 5 classes which has the most improvement on v-mAP@0.5 on our proposed semi-supervised approach compared to the supervised counterpart on UCF101-24 dataset. }
    \label{fig:ucf_vmap_fmap_best}
\end{figure*}

\begin{table*}[t!]
  \centering
  
  \begin{tabular}{ cc  cc  }
    \hline
    \multicolumn{4}{c}{Strong Augmentations} \\
    \hline
    Type & Probability & Random Value & Explanation \\
    \hline
    \hline
    Contrast & 0.7 & 0.8 & Random uniform selection between [0.6, 1.4) \\
    Hue & 0.7 & 0.05 & Random uniform selection between [-0.1, 0.1)\\
    Brightness & 0.7 & 0.9 & Random uniform selection between [0.6, 1.4) \\
    Saturation & 0.7 & 0.7 & Random uniform selection between [0.6, 1.4)\\
    Grayscale & 0.6 & - & -\\
    Gaussian Blur & 0.5 & $\sigma_{x}$=0.1, $\sigma_{y}$=2.0 & Kernel size=(3, 3) \\
    \hline
    \multicolumn{4}{c}{Weak + Strong Augmentation}\\
    \hline
    \hline
    Horizontal Flip & 0.5 & - & -\\
    \hline
  \end{tabular}
  \caption{ \textbf{Details} about selection of random parameters for spatial augmentations.
  }
  \label{tab:aug_details}
\end{table*}

\begin{figure*}[t!]
\begin{center}
\includegraphics[width=\linewidth]{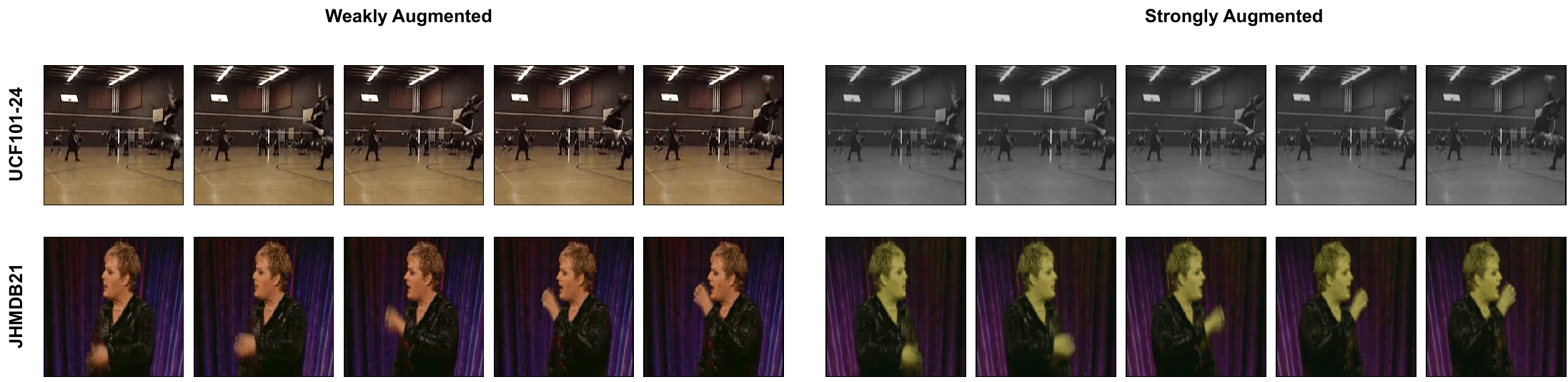}
\end{center}
\caption{\textbf{Visualization of augmentations:} This figure shows the original clip and augmented clip from UCF101 and JHMDB21 dataset respectively. }
\label{fig:spatial_aug}
\end{figure*}

\paragraph{Classwise Performance Analysis}

In this study, we deep diver into the f-mAP and v-mAP of different classes. Here, we discuss performance at specific threshold of 0.5. From the figures \ref{fig:jhmdb_vmap_fmap_best}, we show that the classes with the most improvement with our Stable Mean Teacher approach. The classes with improvement on f-mAP@0.5 and v-mAP@0.5 are \texttt{brush\_hair, kick\_ball, sit, walk, wave} and \texttt{brush\_hair, jump, sit, throw, walk} respectively. Some of these classes have very fast motion. Most improvement on those classes shows that our approach is more robust to motion changes and the predictions are more temporally coherent.

We extend this analysis to UCF101-24 dataset as well. From Fig. \ref{fig:ucf_vmap_fmap_best}, classes with the most gain are \texttt{CliffDiving, Diving, HorseRiding, Skijet, Surfing}, and, \texttt{Diving, Skijet, Skateboarding, Surfing, DogWalking}. A major boost in \texttt{Diving} and \texttt{Surfing} also corroborates our claim that Stable Mean Teacher is less susceptible large motion changes and also small objects.

\begin{figure*}[t!]
\begin{center}
\includegraphics[width=\linewidth]{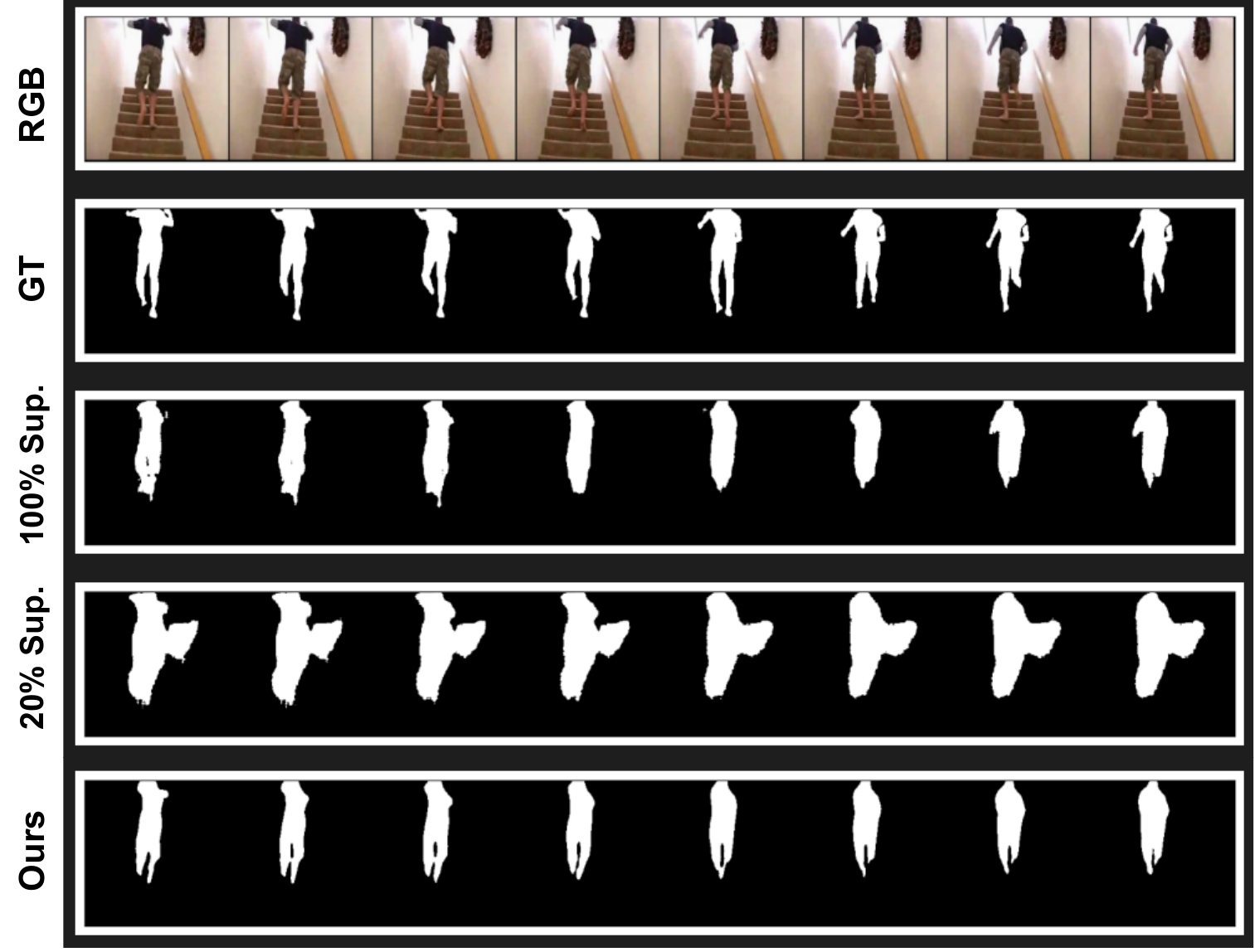}
\end{center}
\caption{\textbf{Qualitative results - Case - I - Boundary Refinement
}  In this scenario, we can see the Ours could even separate out the instance of two legs separately which shows that the precise error from EoR model helps in refinement for fine-grained details. The predictions are even better than the 100\% supervised model.
}
\label{fig:fig1}
\end{figure*}

\begin{figure*}[t!]
\begin{center}
\includegraphics[width=\linewidth]{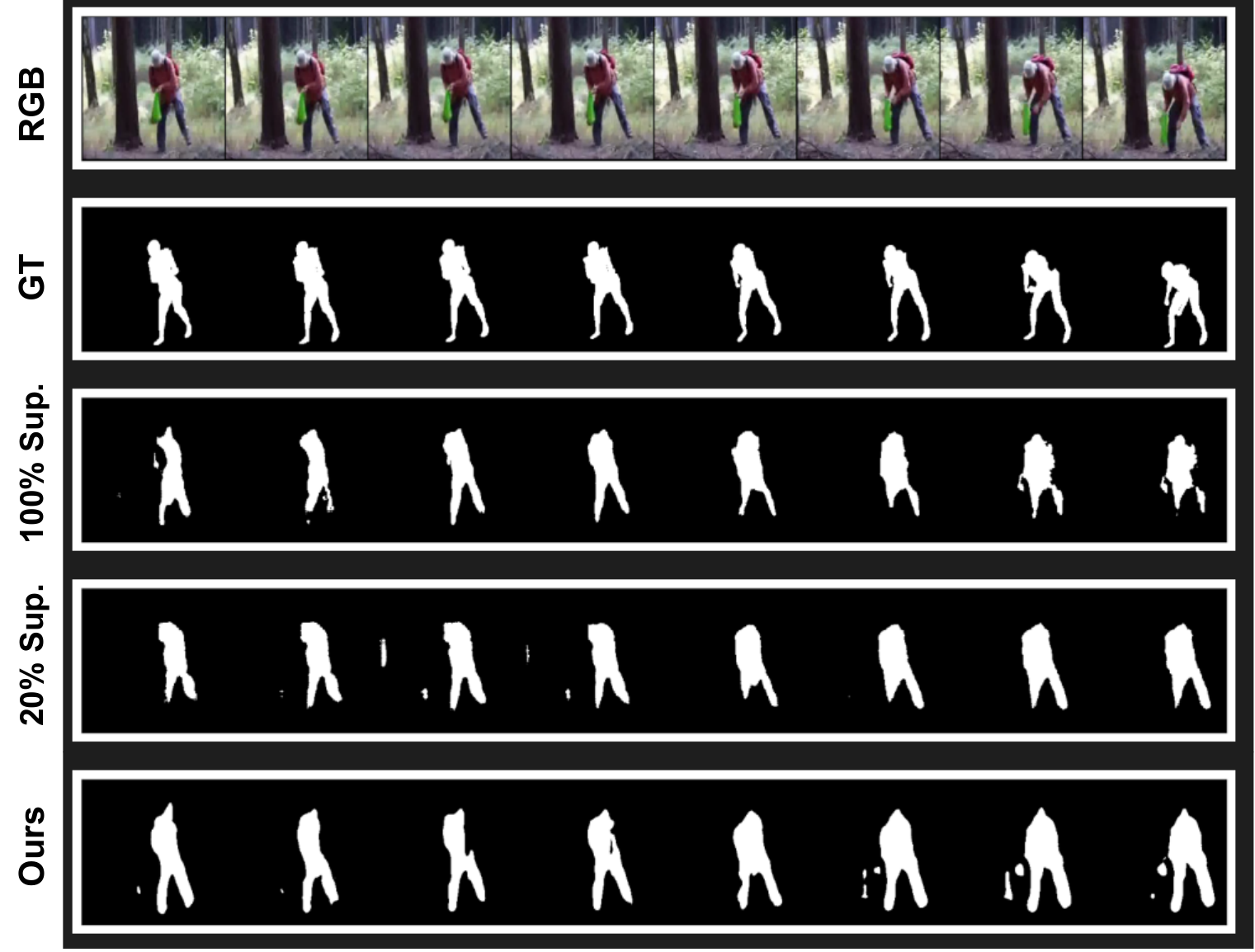}
\end{center}
\caption{\textbf{Qualitative results - Case - II - Noise Suppression
}  In this scenario, we can see the Ours is able to suppress the background noise more better. The detaching of EoR module from main model helps this procedure. Otherwise, the mispredictions gets enhanced.
}
\label{fig:fig2}
\end{figure*}

\begin{figure*}[t!]
\begin{center}
\includegraphics[width=\linewidth]{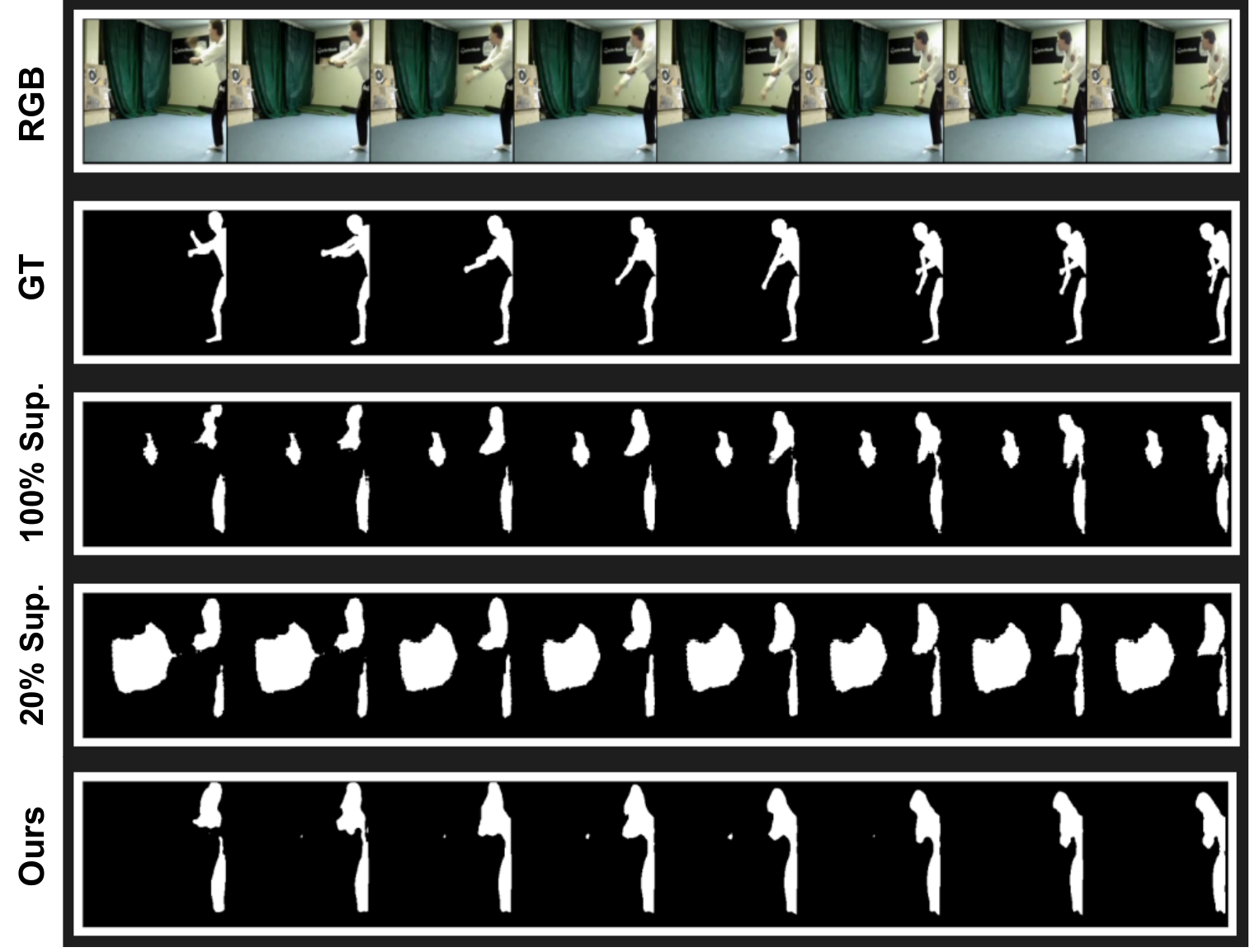}
\end{center}
\caption{\textbf{Qualitative results - Case - III - Noise Suppression + Boundary Refinement
}  In this scenario, model is able to do both getting rid of noise and refining the boundary at the same time. Even 100\% supervised model fails at it.  
}
\label{fig:fig3}
\end{figure*}

\begin{figure*}[t!]
\begin{center}
\includegraphics[width=\linewidth]{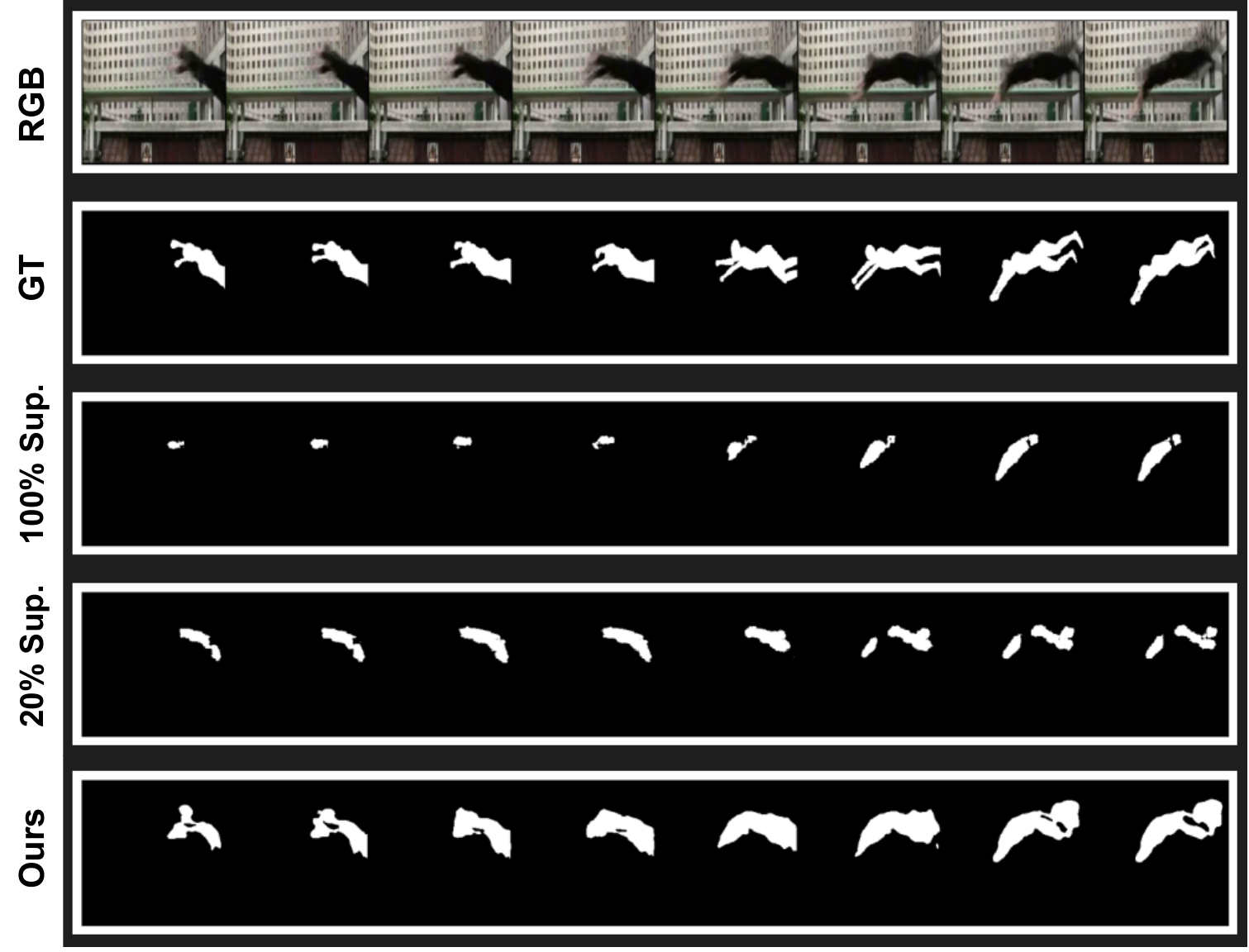}
\end{center}
\caption{\textbf{Qualitative results - Case - IV - Temporal Mask Coherency
}  In this scenario, we show that Ours not only helps to localize the actor spatially but the temporal coherency of mask is also maintained in case of large displacement/motion.}
\label{fig:fig4}
\end{figure*}

\section{Implementation Details}
\label{sec:implement}

We go through architecture, data augmentation and training details in depth here.

\subsection{EoR Architecture} 

In our work, we use a modified version of UNet 3D architecture. It is a simple extension of it's 2D version where 2D Convolution block is replaced by a 3D convolution block and the upsample mode is \textit{trilinear} instead of \textit{bilinear}. Original UNet 3D model has a lot of trainable parameters and to reduce the extra overhead of trainable parameters, we reduce the depth in our EoR Module. Original model has 5 channels depth, and the variation in depth goes like this, $32 \rightarrow 64 \rightarrow 128 \rightarrow 256 \rightarrow 512 \rightarrow 256 \rightarrow 128 \rightarrow 64 \rightarrow 32$. In our case, we reduce the number of channels. EoR architecture have this variation, $16 \rightarrow 32 \rightarrow 64 \rightarrow 128 \rightarrow 64 \rightarrow 32 \rightarrow 16$. This brings down the number of trainable parameters to approximately 1.1M. We also changed the EoR model with various depth and compared the performance. 

\subsection{Data Augmentation Details}

We study both spatial and temporal augmentations to generate weak and strong views. First, video is passed through temporal augmenter block, which temporally augments the video frames. After temporal augmentation, the video passes through the spatial augmenter block. Augmenting in this sequence makes the process \textit{computationally efficient} as for spatial augmentation we only perform augmentation of required frames instead of augmenting all the video frames. The strong augmentation includes random crop, gaussian blur, horizontal flip, grayscale, hue, saturation, brightness, and contrast, whereas, the weak augmentation includes random crop and horizontal flip.

We break into two: 1) Spatial: For a weak augmented view, only random horizontal flip is applied. On the other hand, for a strong view, all types of augmentations are applied.  To measure the consistency between student and teacher prediction, if the frames of weak augmented video is flipped, then, correspondingly frames of strong augmented video is flipped, i.e., geometrical transformation is maintained. 2) Temporal: This augmentation is also similar for both teacher and student. This is because we need to calculate localization consistency across each frame.
 We determine this hyperparameter empirically. One out of three temporal augmentations is chosen  randomly equal probability.

\subsection{Training Details} 
Some extra training details:  The weight is gradually ramped up till 15 epochs and kept consistent after that. 
We use Adam optimizer with an initial learning rate set to 0.0001.  Next, we discuss the augmenter setup to generate the two augmented views. In our work, we use different set of augmentations for weak and strong augmented view. For weak, we use Random Horizontal Flip, and, for strong, we use Color Jitter, Grayscale and Gaussian Blur. In Table \ref{tab:aug_details}, we show the random probability parameters with which these are augmentations are applied on the video. We have not search for best hyperparameter settings for data augmentation in our work. In Fig. \ref{fig:spatial_aug}, an example of spatial augmentation only is shown for two videos, one from UCF101 and one from JHMDB21.

\section{Qualitative Analysis}
\label{sec:qualitative}

Figures \ref{fig:fig1} - \ref{fig:fig4} show couple of more examples showing analyzing the model output qualitatively between different settings. In all figures, GT means ground truth. Successive rows shows predicted localization maps for 100\% and 20\% fully supervised trained model and Ours.

\bibliography{aaai25}

\end{document}